\documentclass{article}

\usepackage{abstract}  % 
\usepackage{PRIMEarxiv}

\usepackage[utf8]{inputenc} % allow utf-8 input
\usepackage[T1]{fontenc}    % use 8-bit T1 fonts
\usepackage{hyperref}       % hyperlinks
\usepackage{url}            % simple URL typesetting
\usepackage{booktabs}       % professional-quality tables
\usepackage{amsfonts}       % blackboard math symbols
\usepackage{nicefrac}       % compact symbols for 1/2, etc.
\usepackage{microtype}      % microtypography
\usepackage{lipsum}
\usepackage{fancyhdr}       % header
\usepackage{graphicx}       % graphics
\graphicspath{{media/}}     % organize your images and other figures under media/ folder

\usepackage{amsmath}
\usepackage{subcaption}

\usepackage{setspace}

\usepackage[utf8]{inputenc}
\usepackage[T1]{fontenc}

\title{JEMA: A \textbf{J}oint \textbf{E}mbedding Framework for Scalable Co-Learning with \textbf{M}ultimodal \textbf{A}lignment
}

\author{
  João Sousa \\
  Faculty of Engineering of the University of Porto (FEUP), Porto, Portugal \\
  Artificial Intelligence and Computer Science Laboratory (LIACC), FEUP, Porto, Portugal \\
  Associate Laboratory for Energy, Transports and Aerospace (LAETA), INEGI, Porto, Portugal \\
  Fraunhofer Institute for Material and Beam Technology IWS, Dresden, Germany \\
  \texttt{jpsousa@inegi.up.pt} \\
    \AND
  Roya Darabi \\
  Faculty of Engineering of the University of Porto (FEUP), Porto, Portugal \\
  Associate Laboratory for Energy, Transports and Aerospace (LAETA), FEUP, Porto, Portugal \\
  \texttt{rdarabi@inegi.up.pt} \\
   \And
  Armando Sousa \\
  Faculty of Engineering of the University of Porto (FEUP), Porto, Portugal \\
  Centre for Robotics in Industry and Intelligent Systems, INESC TEC, Porto, Portugal \\
  \texttt{asousa@fe.up.pt} \\
   \And
  Frank Brueckner \\
  Fraunhofer Institute for Material and Beam Technology IWS, Dresden, Germany \\
  Department of Engineering Sciences and Mathematics, Luleå University of Technology, Luleå, Sweden \\
  \texttt{frank.brueckner@iws.fraunhofer.de} \\
   \And
  Luís Paulo Reis \\
  Faculty of Engineering of the University of Porto (FEUP), Porto, Portugal \\
  Artificial Intelligence and Computer Science Laboratory (LIACC), FEUP, Porto, Portugal \\
  \texttt{lpreis@fe.up.pt} \\
   \And
  Ana Reis \\
  Faculty of Engineering of the University of Porto (FEUP), Porto, Portugal \\
  Associate Laboratory for Energy, Transports and Aerospace (LAETA), FEUP, Porto, Portugal \\
  \texttt{arlr@fe.up.pt} \\
}

\begin{document}
\maketitle
%\newpage
\vspace{50pt}  % Space before the line
\noindent
\vspace{5pt}  % Space after the line
\begin{abstract}
This work introduces JEMA (Joint Embedding with Multimodal Alignment), a novel co-learning framework tailored for laser metal deposition (LMD), a pivotal process in metal additive manufacturing. As Industry 5.0 gains traction in industrial applications, efficient process monitoring becomes increasingly crucial. However, limited data and the opaque nature of AI present challenges for its application in an industrial setting. JEMA addresses this challenges by leveraging multimodal data, including multi-view images and metadata such as process parameters, to learn transferable semantic representations. By applying a supervised contrastive loss function, JEMA enables robust learning and subsequent process monitoring using only the primary modality, simplifying hardware requirements and computational overhead.
We investigate the effectiveness of JEMA in LMD process monitoring, focusing specifically on its generalization to downstream tasks such as melt pool geometry prediction, achieved without extensive fine-tuning. 
Our empirical evaluation demonstrates the high scalability and performance of JEMA, particularly when combined with Vision Transformer models. We report an 8\% increase in performance in multimodal settings and a 1\% improvement in unimodal settings compared to supervised contrastive learning. Additionally, the learned embedding representation enables the prediction of metadata, enhancing interpretability and making possible the assessment of the added metadata's contributions.
Our framework lays the foundation for integrating multisensor data with metadata, enabling diverse downstream tasks within the LMD domain and beyond.
\end{abstract}

% keywords can be removed

\section{Introduction}

% high level
%\IEEEPARstart{T}{he} 
The application of Machine Learning (ML) in manufacturing processes can significantly contribute to the wider adoption of advanced manufacturing technologies by simplifying complex tasks and reducing the need for highly specialized machine operators or process engineers \cite{Dogan2021Machine, 10216797}. This is particularly relevant in today's manufacturing landscape, caused by rapid technological advancements, an aging population, socio-economic conditions, external forces, job attractiveness, job satisfaction, and industry limitations, which are leading to a shortage of skilled workers \cite{Akomah2020Skilled, Juricic2021Review}.

Additive manufacturing processes (AM) are transforming the manufacturing industry because they offer unprecedented flexibility in part geometry and material usage \cite{7740609}. In particular, Laser Metal Deposition (LMD), also known as Directed Energy Deposition (DED) or Laser Cladding, is a metal AM process capable of building, repairing, and coating parts with high energy and material efficiency compared to traditional processes such as welding or machining \cite{Karayel2020Additive, met11040672}. Metal-AM processes are complex dynamic systems involving multiple physics with several input parameters \cite{guan_modeling_2020, SVETLIZKY2021271}. To manage this complexity, monitoring systems are often employed to fine-tune the process or detect anomalies, thereby preventing defects in manufactured parts. In LMD, we can monitor the process with several types of sensors: image-based sensors (visible or thermal cameras), pyrometers, and audio or machine data (e.g., process input parameters) \cite{tang_review_2020, Simon_J}. Previous works often report sensor data fusion. When only limited datasets are used, data fusion is normally achieved by pre-processing raw sensor data \cite{MOZAFFAR2022117485, JAMNIKAR202342}.

% Multimodal in geral systems/nature

The world exhibits an intrinsically multi-physics, thus, multimodal nature. Multimodal data refers to heterogeneous data or cognitive signals describing various aspects of the same object, recorded in different kinds of media such as text, images, videos, or sound. The term 'modality' denotes a particular way or mechanism of encoding information \cite{8715409}. Multimodal data can depict an object from different attributes or viewpoints, typically complementing the content of unimodal data \cite{huang2021makes}. 

Multimodal machine learning or multimodal AI (MAI) aims to construct models capable of processing and correlating information from multiple modalities. The primary objective of multimodal representation learning is narrowing the distribution gap in a joint semantic subspace while keeping modality-specific semantics intact. One way to bridge the heterogeneity gap of different modalities is through joint representation, where unimodal representations are projected into a shared semantic subspace for feature fusion \cite{8269806}. This cross-modal similarity method focuses on learning coordinated representations while preserving inter-modality and intra-modality similarity structures, aiming to minimize distances between similar semantics or objects and maximize distances between dissimilar ones \cite{10041115, 9952528}. MAI has demonstrated superior potential compared with unimodal methods on several files such as medical \cite{7827160}, education \cite{lee2023multimodality}, robotics \cite{10275168} and emotion recognition \cite{10213035}.

% Our problem and approach

\begin{figure*}[h!]
\centering
\includegraphics[width=1.0\textwidth]{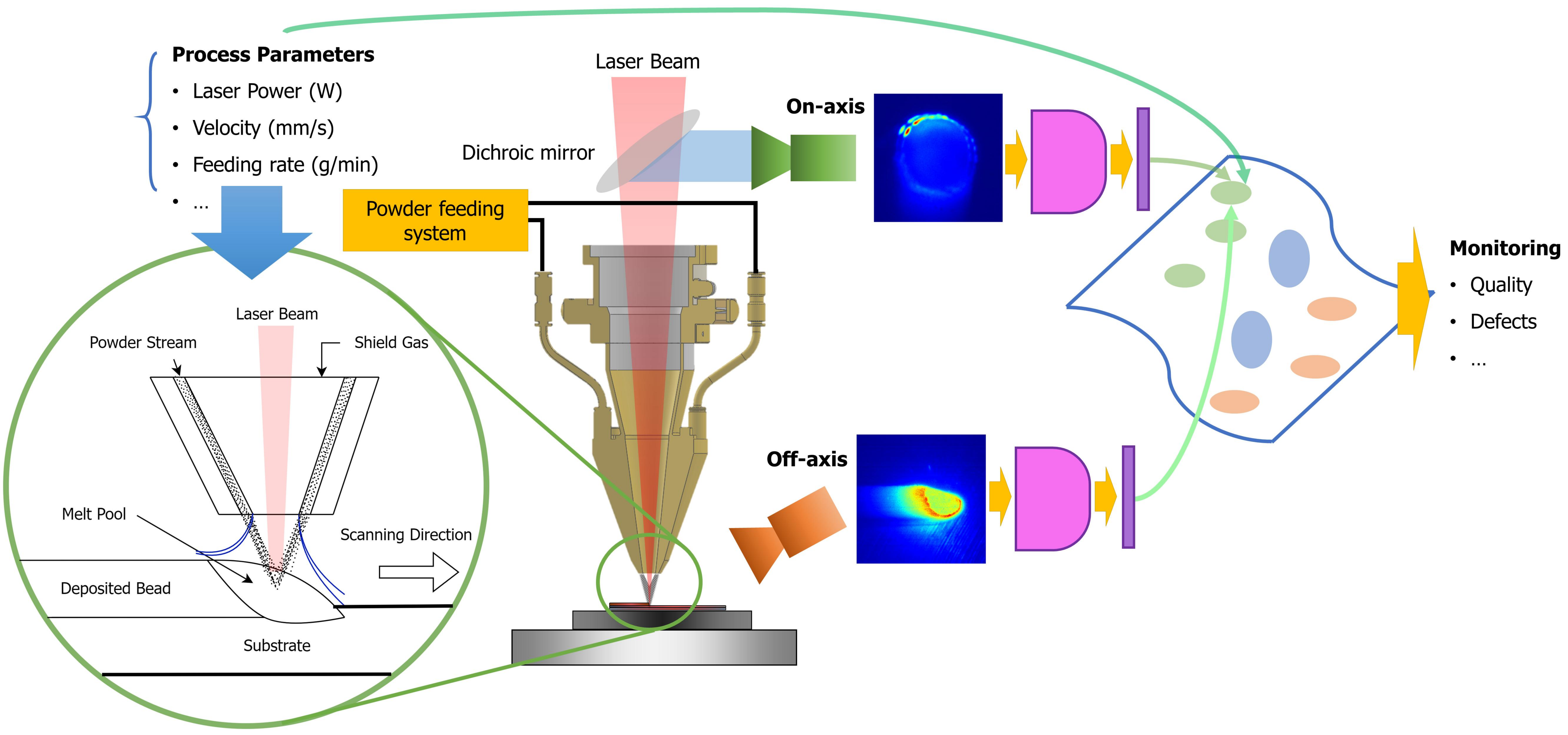}
\caption{Diagram of a Laser Metal Deposition (LMD) system with an on-axis and off-axis monitoring solution. It shows the influence of process parameters for multi-modality representation for monitoring LMD.}\label{ded_jema}
\end{figure*}

This work focuses on the application of MAI in an industrial process, particularly in LMD. LMD has three main process parameters: laser power ($P$, $W$), scanning feed rate ($v$, $\text{mm s}^{-1}$), and powder mass rate ($\dot{m}$, $\text{g min}^{-1}$) \cite{Mahamood2018Processing}. In our setup, Figure~\ref{ded_jema}, we have developed a monitoring solution based on the Robot Operating System (ROS 2) \cite{ros2} to leverage ROS 2 capabilities in standardization, modularity, interoperability, and multimodal capabilities. Thus, we collected data from: (i) process parameters - $P$ and $v$; (ii) the on-axis visible light camera; and (iii) the off-axis thermal camera. Based on the collected data in the single-track printing experiments, our goal is to transfer knowledge to the main modality: on-axis data. This modality allows for more efficient and easier process monitoring due to its minimal variation in printing direction and offering a direct view of the laser-material interaction. 
Additionally, it provides real-time feedback, which will be critical for implementing closed-loop control in future work.
%Additionally, the on-axis data provides real-time feedback, which will be crucial for implementing closed-loop control of the deposition process in future work. 

%%% IMAGEM %%%%%

% IMAGEM DO SETUP 
%\begin{figure*}[!t]
%\centering
%\subfloat[]{\includegraphics[width=2.5in]{fig1}%
%\label{fig_first_case}}
%\hfil
%\subfloat[]{\includegraphics[width=2.5in]{fig1}%
%\label{fig_second_case}}
%\caption{Dae. Ad quatur autat ut porepel itemoles dolor autem fuga. Bus quia con nessunti as remo di quatus non perum que nimus. (a) Case I. (b) Case II.}
%\label{fig_sim}
%\end{figure*}

In this work we have combined the Vision Transformer model (ViT) with a novel supervised contrastive loss to embed process parameter information to tackle the following problems: (i) representation: how to create an embedding representation from the two cameras; (ii) Alignment: based on the process parameters, how we can add context to our representations to learn better representations by capturing cross-modal interactions between process parameters.; and (iii) Transference - how to leverage knowledge gained from multiple data sources (process parameters, on-axis camera, and off-axis camera) to improve process monitoring using only the primary modality. Our approach achieved state-of-the-art results for the regression task. Furthermore, using JEMA we can perform inference on the embedding representation to contextualize the predictions in both multi-modal and uni-modal settings.

% paper structure 
The remainder of this paper is structured as follows: Section II provides a review of relevant studies. Section III describes the problem formulation and the baseline techniques employed. Section IV presents the proposed method and the similarity loss functions for JEMA. Section V showcases the results, comparing them with baseline methods. In Section VI, a comprehensive discussion on the main advantages of the proposed method is presented, along with an acknowledgment of its limitations and potential drawbacks. Finally, Section VII concludes the article by summarizing the key findings of the research and outlining potential directions for future research, indicating how this work could be built upon to further advance the field."

\section{Related Work}

In this section, we review seminal work in monitoring LMD, image transformers, and Multimodal Learning. Moreover, we dive into representation learning, alignment and co-learning to contextualize the proposed approach.

\subsection{LMD monitoring} 

Monitoring in LMD is crucial for ensuring quality and detecting defects during production, minimizing waste, improving efficiency, and potentially leading to real-time adjustments. Integrating AI with real-time sensor data can enhance monitoring by processing diverse data sources, detecting patterns, and making predictions \cite{MOZAFFAR2022117485}. As LMD has a multi-physics nature, multimodal data is frequently used in a data fusion approach. For example, Perani et al. \cite{PERANI2023102445} combined on-axis images and process parameters based on a custom Convolutional Neural Network (CNN) to process the image and add an Artificial Neural Network (ANN) to combine image features and process parameters to predict the geometry of the deposition. In a similar approach, Jamnikar et al. \cite{jamnikar_-process_2022, jamnikar_comprehensive_2021-1, jamnikar_situ_2022} combined image with pyrometer data to predict process parameters, geometry of the deposition, and material properties. Tian et al.\cite{tian_deep_2021} developed a method to predict the quality of the deposition (good or bad) based on a weighted average prediction of two different models: a CNN to process image data from a pyrometer (PyroNet); and, a Long-term Recurrent Convolutional Network (LRCN) to process sequences of infrared image data (IRNet). The same team \cite{mcgowan_physics-informed_2022} evaluated fusing data from real and virtual data using physics-informed loss functions but their effectiveness in identifying porosity defects was not substantial \cite{mcgowan_physics-informed_2022}. With small datasets, data fusion is used with pre-processed data and combined with an ML model such as LSTM \cite{perani_long-short_2023}, ANN \cite{kim_infrared_2023} or others shallow approaches \cite{ye_predictions_2022, chen_multisensor_2023}. Although some authors tried to evaluate the data representations by visualizing the embeddings using t-sne \cite{PANDIYAN20221064}, transference between modalities is still not explored. 

Pandiyan et al. \cite{PANDIYAN20221064} used a semi-supervised method based on triplet contrasting loss to classify six process qualities, produced by different process parameters. More recently \cite{pandiyan_real-time_2023}, with a similar dataset achieved better results using a self-supervised learning approach with Bootstrap your own latent (BYOL) \cite{grill2020bootstrap}.

\subsection{Image Transformers}

Deep learning, particularly Convolutional Neural Networks (CNNs), has revolutionized computer vision tasks such as image classification \cite{8379889}, segmentation \cite{9848977}, and object detection \cite{8650517}. However, CNNs face challenges when dealing with large-scale and highly complex datasets. This paved the way for vision transformers, a new architecture that adapts the successful transformer model, originally prevalent in Natural Language Processing (NLP), for image analysis \cite{10410015}. The ViT model, proposed by Dosovitskiy et al. \cite{dosovitskiy2021image}, was designed to overcome the limitations of CNNs in capturing long-range relationships and global information in images. ViT relies on a series of transformer blocks, each combining a multi-head self-attention mechanism with a feedforward neural network.
ViTs and their variants \cite{10410015} have demonstrated potential in various domains including medical image analysis \cite{10416046, 10474871}, robotics \cite{liu2023vita, goyal2023rvt}, remote sensing \cite{9883054}, and pose estimation \cite{9956315}.

%Let \( X \) be the input image, divided into \( N \) patches, each of dimension \( D \times D \). The linear embedding of each patch \( x_i \) is represented as \( x_i = W_0 \cdot \text{patch}_i + b_0 \), where \( W_0 \) is the learnable weight matrix and \( b_0 \) is the learnable bias term. These embedded patches are then tokenized and linearly projected to obtain the input sequence \( X' = \{x_1', x_2', ..., x_N'\} \), where \( x_i' = W_q \cdot x_i + b_q \). The sequence \( X' \) is then processed through \( L \) transformer blocks. Each block consists of multi-head self-attention followed by a feedforward neural network. The output of the final transformer block is passed through a linear layer to obtain the classification scores or regression outputs.

Drawing inspiration from how our brains process information from vision, sound, and language, researchers are increasingly turning to Transformers as a powerful tool for multimodal learning \cite{Khan_2022}. Specifically, a Transformer's input can consist of one or multiple token sequences, each with its own attributes like modality labels or sequential order. This inherent flexibility makes Transformers particularly well-suited for multimodal AI (MAI), as they can handle various data types without the need for specialized architectures, unlike traditional methods \cite{10123038}.

\subsection{Multimodal Learning}
% Intro to multimodal Learning and SoA

Multimodal Learning has been used in several fields, including medical research \cite{truhn_large_2024}, music generation \cite{Kang2024}, autonomous driving \cite{teshima_determining_2024, wu_joint_2024}, and environmental protection \cite{Shadrin2024}. Based on self-supervised learning for short or few-shot learning, and transformer models in NLP, MAI gained popularity with models such as CLIP \cite{radford2021learning}, which combines text with images in a supervised contrastive approach, as well as BEiT \cite{wang2022image} and Flamingo \cite{alayrac2022flamingo}. For more general approaches, the Gato model \cite{reed2022generalist} is a multimodal and multi-task framework capable of playing Atari games, captioning images, engaging in chat, stacking blocks with a real robot arm, and more, deciding based on context whether to output text, joint torques, button presses, or other tokens. Facing a similar challenge of building a unified network, Meta-Transformer \cite{zhang2023metatransformer} can unify 12 modalities with unpaired data for tasks such as perception (text, image, point cloud, audio, video), practical applications (X-Ray, infrared, hyperspectral, and IMU), and data mining (graph, tabular, and time-series).

\subsection{Representation Learning}

% Intro to Contrastive Learning and SoA see if can merge with the previous one

Although supervised deep learning has revolutionized image processing, it has necessitated the building of specialist models for individual tasks and application scenarios. Applying this approach to smaller datasets or situations with limited labeled data remains challenging \cite{10423573}. Self-supervised representation learning methods promise a single universal model that would benefit a wide variety of tasks and domains. These methods have shown success in natural language processing and computer vision domains, achieving new levels of performance while reducing the number of labels required for many downstream scenarios. Vision representation learning is experiencing similar progress in three main categories: generative, contrastive, and predictive methods \cite{e26030252, liang2023foundations, 9893562}. However, this work excludes from consideration generative approaches.

In contrastive learning, the model learns to represent similar data points close together and dissimilar ones far apart. This approach works for both labeled (supervised) and unlabeled data (unsupervised), making it a powerful technique for self-supervised learning (SSL) \cite{Khac2020Contrastive}. In SSL, SimCLR \cite{chen2020simple} utilizes data augmentation in image data to create positive pairs with contrastive learning in a random image batch. Barlow Twins \cite{zbontar2021barlow} achieves a more efficient and informative image representation by training on distorted versions of the same image. It feeds these versions into a network to extract key features and then aligns these features, ensuring they capture the image's essence despite the distortions. In a different approach, BOYOL \cite{grill2020bootstrap} avoids using negative pairs and relies on different neural networks, referred to as online and target networks, which interact and learn from each other.

In predictive methods, data augmentations are not required. For example, I-JEPA \cite{assran2023selfsupervised} learns by predicting the representation of various target blocks of the same image using L2-distance between the predicted patch-level representations. Building on this work, V-JEPA \cite{bardes2024vjepa} is trained similarly using video data, and the L1-distance is applied between target and predicted embeddings. Another non-contrastive method is Variance-Invariance-Covariance Regularization (VICReg) \cite{bardes2022vicreg}, a regularization method that combines the variance term with a decorrelation mechanism based on redundancy reduction and covariance regularization. Although VICReg does not require the embeddings to be identical or even similar, explanation of the prediction representation was not performed.

\subsection{Alignment}

Multimodal alignment involves identifying connections and correlations between components of instances originating from two or more modalities \cite{8269806, liang2023foundations}. Implicit alignment serves as an intermediary, often latent, step for various tasks, leading to enhanced performance across domains such as speech recognition, machine translation, media description, and visual question-answering. These models avoid explicit data alignment and supervised alignment examples; instead, they autonomously learn to align data in latent space (embeddings) during the training phase \cite{8269806}. This is typically achieved using similarity learning in a contrastive approach \cite{duan2022multimodal}. For image and text alignment, Karpathy et al. \cite{karpathy2015deep} employed a dot product similarity between image and description embeddings to generate image descriptions based on a few hard-coded assumptions. Xiong et al. \cite{xiong2016dynamic} enhanced the similarity between text and images using a supervised question-answering approach. Yu et al. \cite{yu2016video} tackled video captioning with a Hierarchical Recurrent Neural Network (RNN) where two RNNs condition the multimodal layer to exploit temporal and spatial attention mechanisms.

\subsection{Co-learning}

Transference, specifically in the context of co-learning, entails transferring knowledge between modalities to support the target modality, which might be affected by noise or resource constraints. Multimodal co-learning involves the exchange of information among modalities, primarily from a resource-rich modality to one with limited resources and noisy inputs, lacking annotated data, and unreliable labels. This process utilizes shared representation spaces between modalities to facilitate effective knowledge transfer \cite{liang2023foundations, 8269806, ZADEH2020188}.

Moon et al. \cite{moon2016multimodal} transfer information from a speech recognition neural network (audio-based) to a lip-reading model (image-based), thereby enhancing visual representation and enabling audio-independent lip-reading during testing. DeViSE \cite{NIPS2013_7cce53cf} improves image classification by establishing a coordinated similarity space between images and text. Marino et al. \cite{marino2017know} utilize knowledge graphs to classify images through a graph-based joint representation. Jia et al. \cite{jia2021scaling} boost image classifiers by employing contrastive representation learning between images and noisy captions.
Arora and Livescu \cite{6639047} enhance acoustic features by applying kernel and canonical correlation analysis on both acoustic and articulatory data. They utilize only articulatory data during construction and depend solely on the resulting acoustic representation during testing.

\section{Preliminaries}

In this section, we introduce the dataset and outline the problem formulation. We also provide a review of previous works, including descriptions of their loss functions.

\subsection{Dataset Definition}

\begin{figure}[h!]%
\centering
\includegraphics[width=0.7\textwidth]{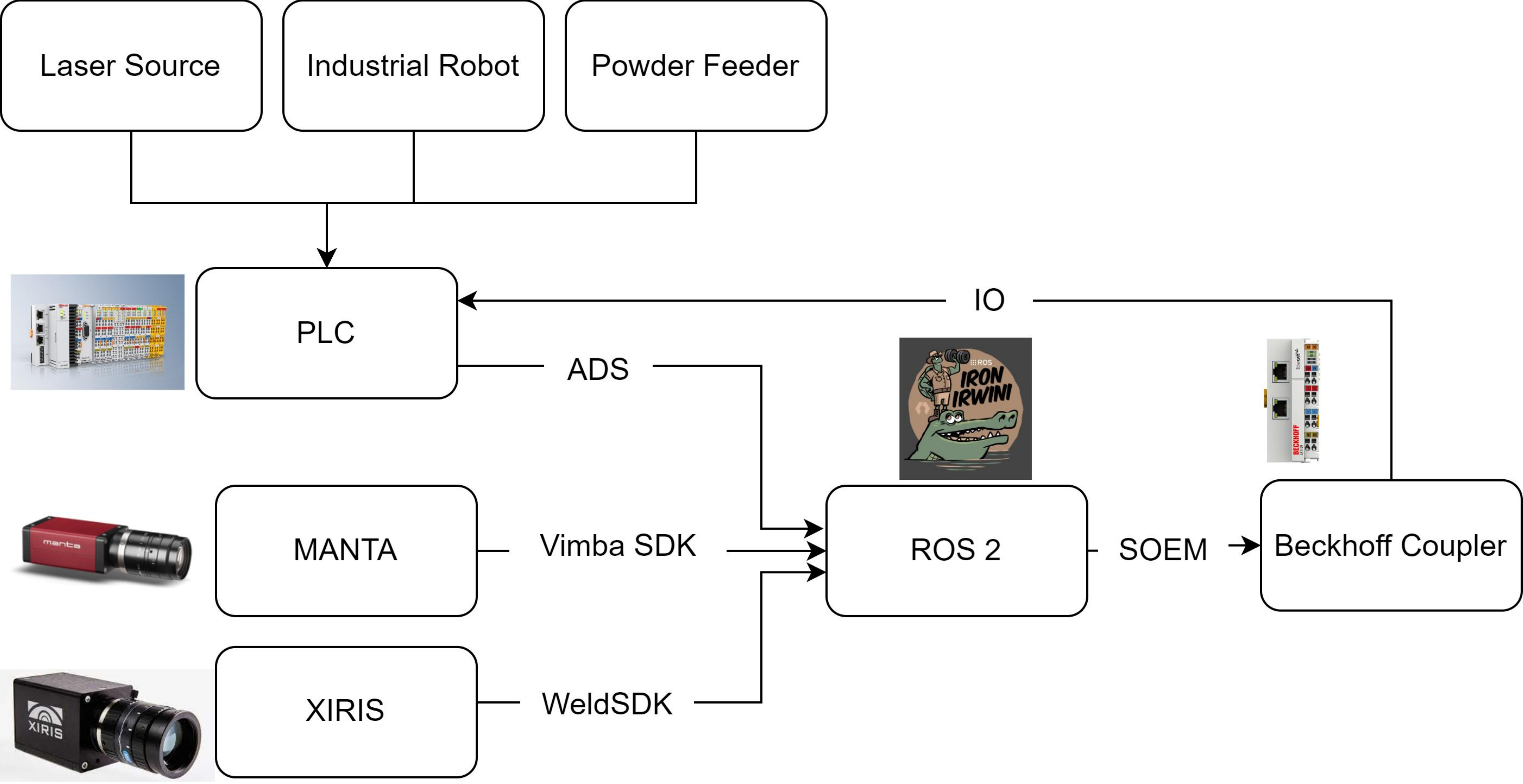}
\caption{Integration of multiple heterogeneous data sources, such as PLC and image-based data, with ROS 2.}\label{ros_inegi2}
\end{figure}

In this study, we used ROS 2 \cite{ROS2_2} to digitize the process by gathering data from multiple devices, as illustrated in Figure~\ref{ros_inegi2}. The industrial robot cell comprises: (i) KUKA KR30 HA robot; (ii) Flowmotion MEDICOAT powder feeding system; (iii) ROFIN FL030 laser with 3kW power and a wavelength of $\lambda = 1064 : nm$; (iv) Fraunhofer IWS COAX12V6 nozzle with a powder splitter; (v) Process Optics D50 from Coherent, emitting a laser beam with a diameter of 1.5 - 3 mm; and (vi) Beckhoff CX8100 PLC that manages all the main devices and safety devices. To complete the loop, Beckhoff coupler EK1100, equipped with digital and analog inputs and outputs, controls the laser power. A ROS node regulates the Beckhoff coupler using the Simple Open EtherCAT Master Library (SOEM). Two sensor cameras are employed: (i) an on-axis visible camera, MANTA G-033B from Allied Vision; and (ii) an off-axis short-wave infrared (SWIR) camera, XIR-1800 from Xiris. Within this framework, we acquired data with softly synchronized timestamps.

%%% IMAGEM %%%%%

\begin{figure*}[h!]
\centering
\subfloat[]{\includegraphics[width=2.0in]{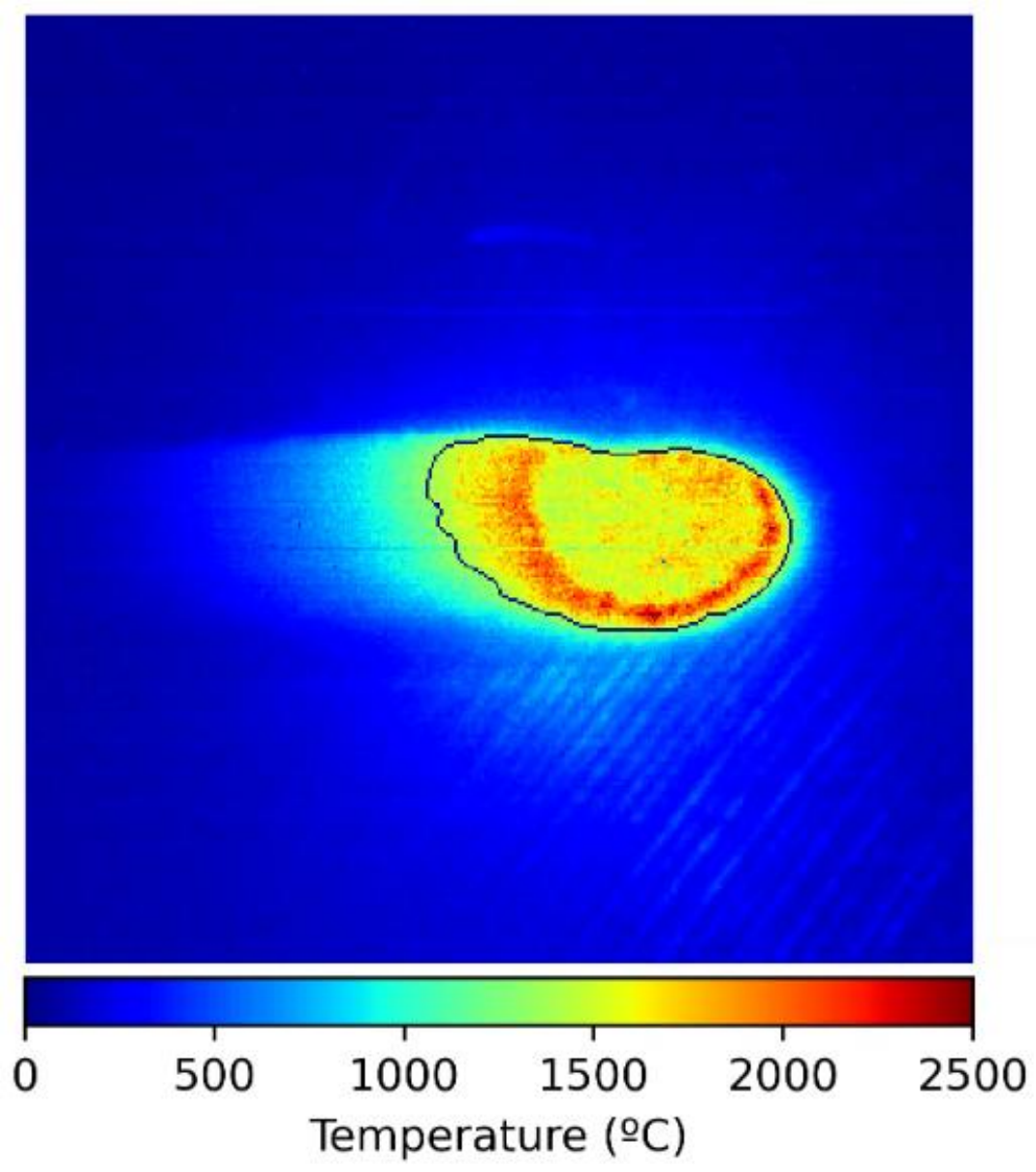}%
\label{xiris_1a}}
\hfil
\subfloat[]{\includegraphics[width=2.0in]{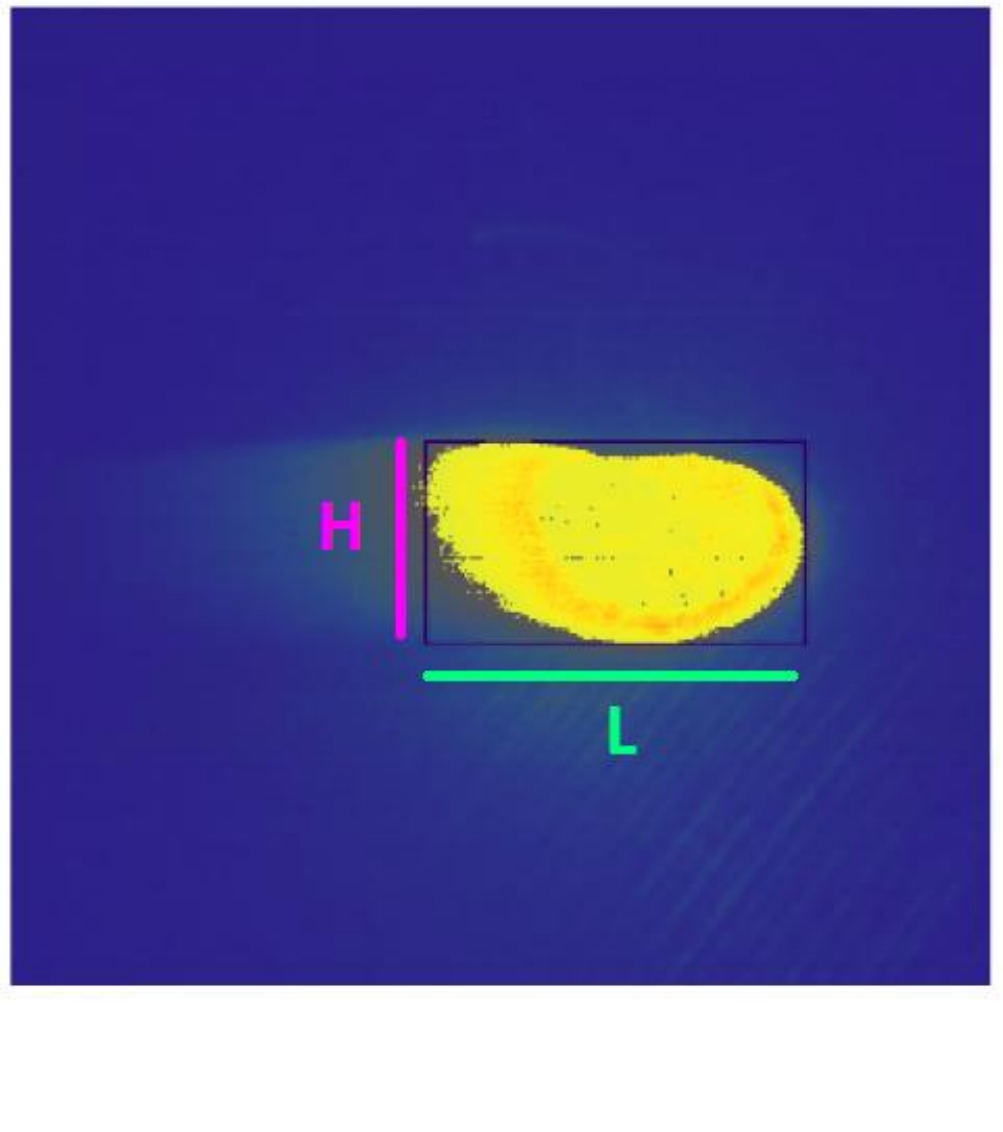}%
\label{xiris_1b}}
\hfil
\subfloat[]{\includegraphics[width=2.0in]{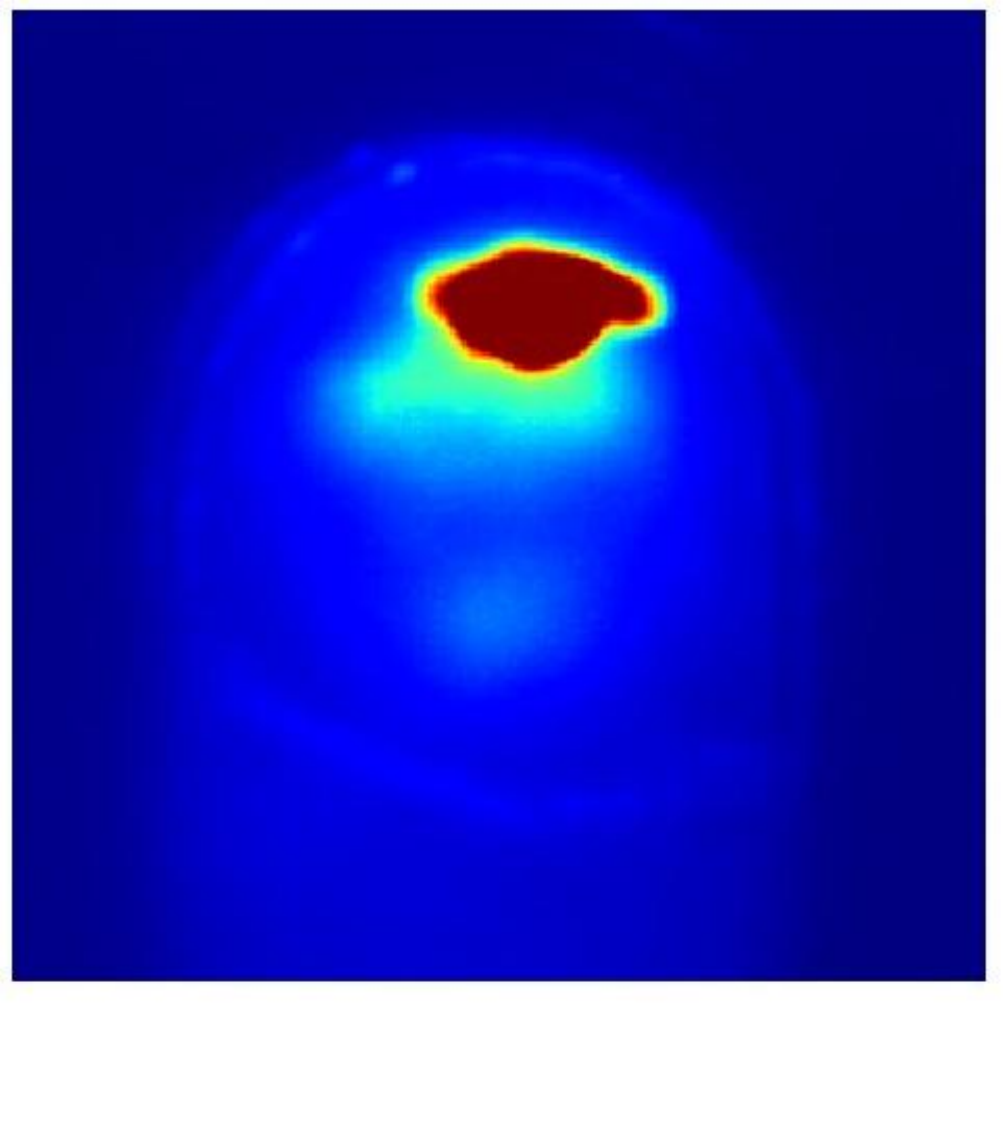}%
\label{manta_1}}
\caption{Images from the same timestamp: (a) on-axis camera with iso-thermal line; (b) definition of melt pool Lenght (L) and Height (H); and (c) off-axis camera.}
\label{fig_xiris}
\end{figure*}

With a Full Factorial DOE with 4 levels of velocity (4, 6, 8, 10 mm$s^{-1}$) and 7 levels of laser power (800, 1000, 1200, 1400, 1600, 1800 and 2000 $W$), we obtained over 27,000 frames, consisting of pair of off-axis and on-axis images, with corresponding process parameters.
In Figure~\ref{xiris_1a} and \ref{xiris_1b}, an off-axis image is shown, where computer vision techniques were employed to identify the isothermal line. This facilitated the masking of the molten pool (MP) and enabled the measurement of its length (L) and height (H). Notably, variations in emissivity caused different material phases (solid - liquid) at the MP's center to appear cooler, posing challenges for precise masking. After applying median filters, a temperature threshold of 1250 °C was selected to represent the MP, as agreed upon by the authors. Additionally, the on-axis image for the same frame is displayed in Figure~\ref{manta_1}.

\begin{figure}[!]
\centering

\subfloat[]{\includegraphics[width=3.0in]{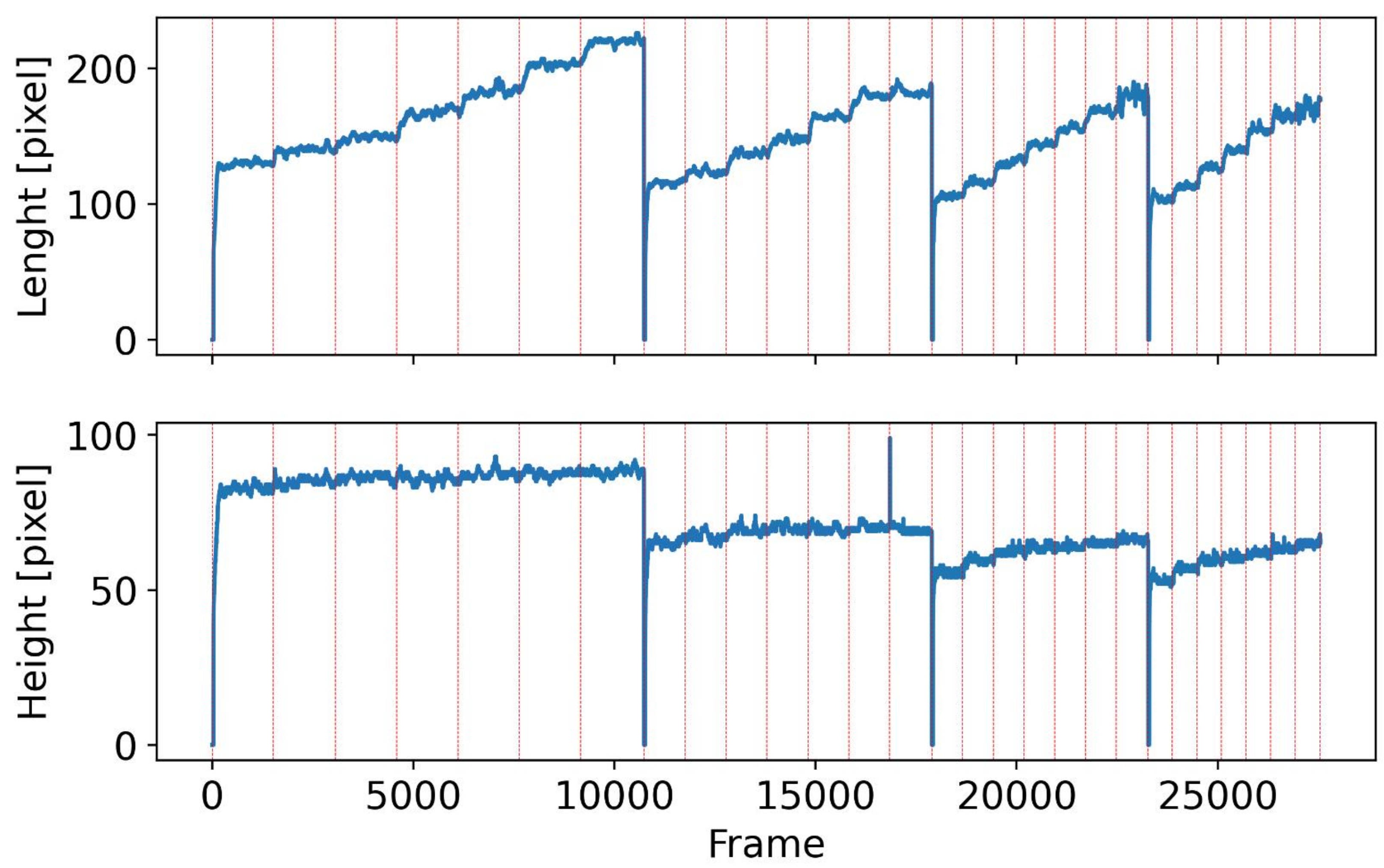}%
\label{DOE_1a}}
\hfil
\subfloat[]{\includegraphics[width=3.0in]{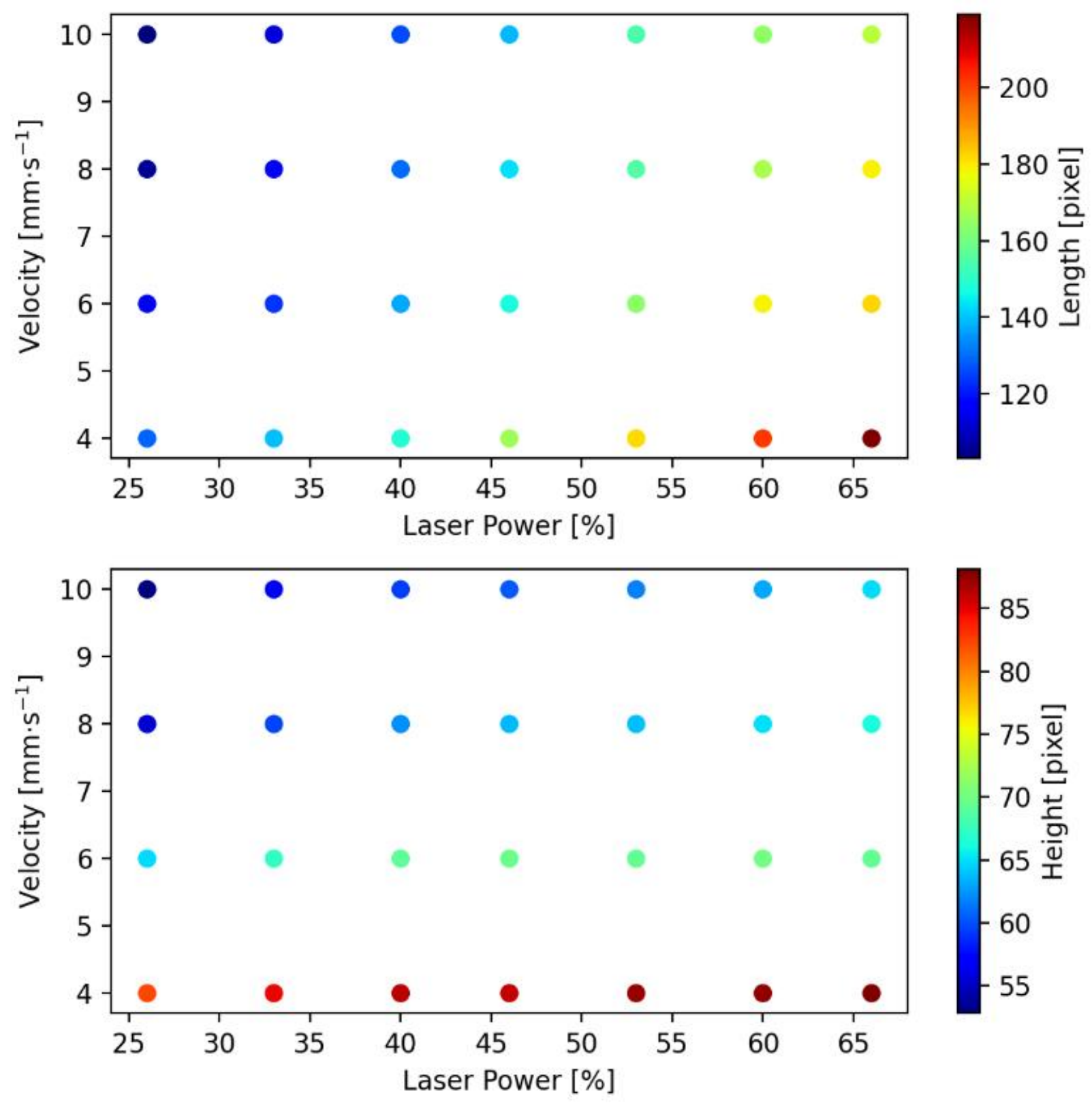}%
\label{DOE_1b}}

\caption{Melt pool lengths (L) and heights (H) in pixels for all design of experiments are presented in (a), while the mean values for each parameter set are shown in (b).}
\label{fig_doe}
\end{figure}

Using this computer vision algorithm, the L and H for all DOE are presented in Figure~\ref{DOE_1a}. For data visualization, we performed a mean of L and H for each set of process parameters as shown in Figure~\ref{DOE_1b}. Outliers with a $\sigma \geq 3.0$ were removed to obtain the final dataset.

% IMAGE Pre Processing

% IMAGE Pre Processing

\subsection{Problem Formulation}
% Problem definition

\begin{figure}[h!]%
\centering
\includegraphics[width=0.45\textwidth]{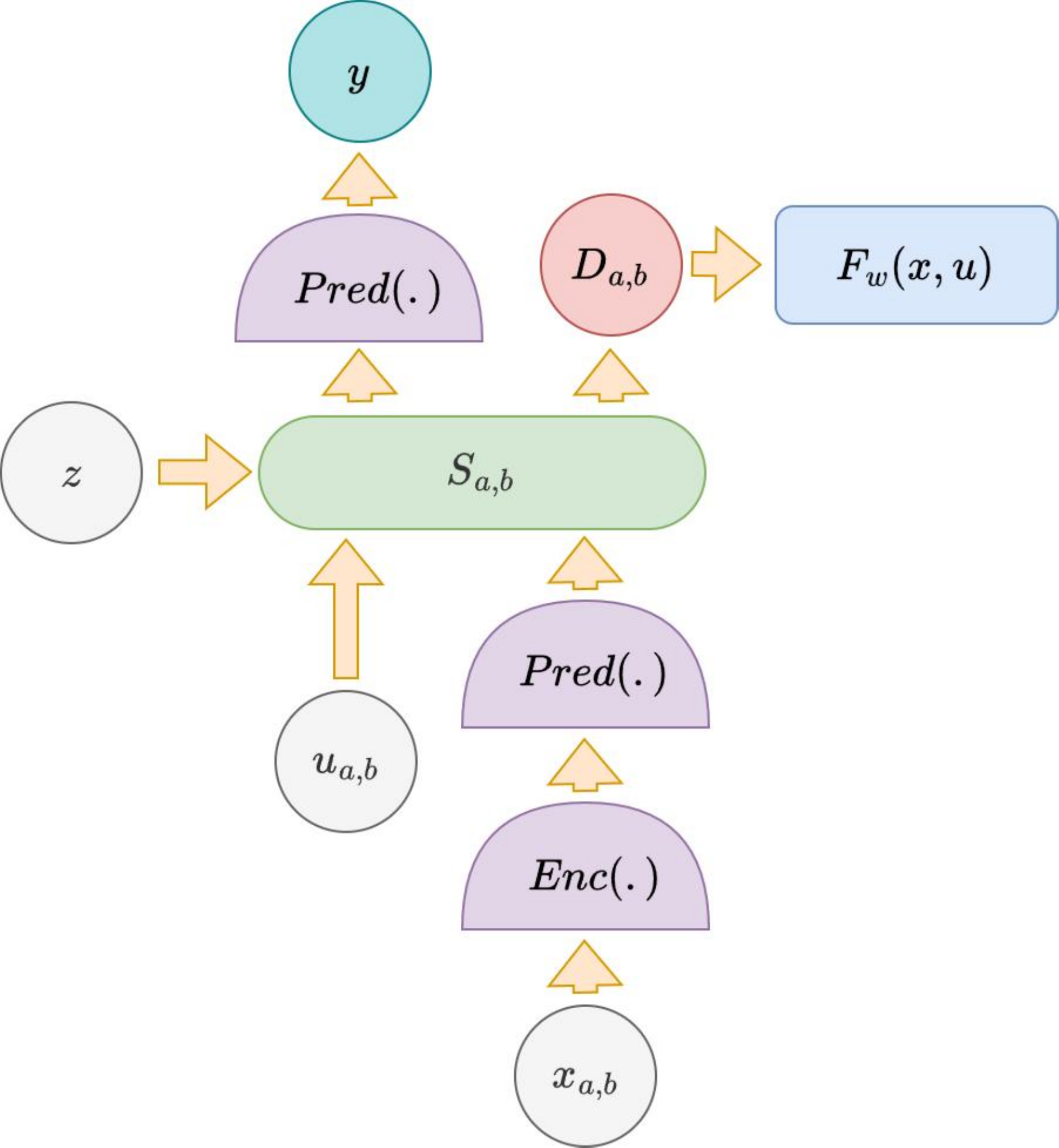}
\caption{Diagram of the components for laser metal deposition monitoring in a multimodal setting. Where $x_{a,b}$ are the multimodal data and $u_{a,b}$ the process parameters that can be use to build a latent representation $S_{a, b}$. }\label{ebm}
\end{figure}

Multimodal Learning encapsulates several challenges, as discussed in Section II. In our problem, we aim to learn a more meaningful embedding space or representation, and improve unimodal predictions (melt pool length and height). First, we consider two image input streams from different modalities: $X_a = { \mathcal{X}_1^n, ..., \mathcal{X}_t^n }$ and $X_b = { \mathcal{X}_1^m, ..., \mathcal{X}_t^m }$, where $\mathcal{X}_t^n$ and $\mathcal{X}_t^m$ represent the $n$-dimensional and $m$-dimensional embeddings of $X_a$ and $X_b$ at time $t$, respectively. Based on process parameters such as laser power $P = { p_1, ..., p_t }$ and velocity $V = { v_1, ..., v_t }$, we can designate the metadata as the inputs in our dynamic system, represented by $U(P,V) = { \mathcal{U}_1, ..., \mathcal{U}_t }$.
To achieve multimodal predictions in a high-dimensional space, Energy-based models (EBMs) can be utilized \cite{dawid2023introduction}. Instead of minimizing the divergence measure between the prediction and the target, we seek $D_{a,b}$'s that satisfy a set of constraints posed $z$, expressed as the energy function $F(x, u)$. Our objective is to assign low energies between modalities, $X$, with the same process parameters, $U$, and regularize representations based on process parameters. Based on the formulation of EBM, we can define it as either $F_w(x, u) = \text{argmin}_z E_w(x, u, z)$ or $F_w(x, u) = -\frac{1}{\beta} \log \int_z \exp(-\beta E_w(x, u, z')) dz'$ as presented schematically in Figure~\ref{ebm}. The inference then is simply $\hat{u} = \text{argmin}_u F_w(x, u)$, where $w$ represents the model weights.

%%% IMAGEM %%%%%

To prevent energy collapse \cite{assran2023selfsupervised}, regularization can be employed. This involves constraining the size of the embedding space and utilizing contrastive methods. Contrastive methods aim to reduce the energy of training samples while simultaneously increasing the energy of appropriately generated contrastive samples that are inherently distinct from the training set. An example of a contrastive loss is the pairwise hinge loss, expressed as follows:
\begin{equation}
\label{eq:hinge_loss}
\begin{split}
\mathcal{L}_{\text{hinge}}(x, u, \hat{u}, w) &= [\max(0, \\
&\quad F_w(x, u) - F_w(x, \hat{u}) + m(u, \hat{u}))]^+
\end{split}
\end{equation}
In this context, $x$ and $u$ represent training data points whose energy is intended to be minimized, while $\hat{u}$ denotes a contrastive point whose energy needs to be elevated. When minimizing $L_{\text{hinge}}$, the objective is to ensure that the energy of training samples remains lower than that of both training and contrastive samples by at least the margin $m$, which is determined by the distance between $u$ and $\hat{u}$. It is essential for suitable contrastive loss functions to maintain a non-zero margin to prevent energy collapse. While contrastive loss functions can be computed pairwise for specific datasets, such as the hinge loss in Equation~\ref{eq:hinge_loss}, recent interest has been growing in batch losses, which operate on batch of training data rather than pairs, such as InfoNCE \cite{oord2019representation, balestriero2023cookbook}.

In addition to the multimodal learning task, we have a regression task aimed at predicting the melt pool lengths ($L$) and heights ($H$). We define our final task as $Y(L,H) = { \mathcal{Y}_1, ..., \mathcal{Y}_t }$. This regression task can be formulated as follows:

\begin{equation}
\label{eq:reg}
%\hat{L}, \hat{H} = f(X_a, X_b)
%E(x, y, w) = \frac{1}{2} || F_w(x) - y||
E(x, y, w) = \frac{1}{2} || F_w(x) - y ||_2^2
\end{equation}

The evaluation of both the representation learning and prediction tasks will be conducted in both multi- and unimodal settings.

\subsection{Attention Mechanism in ViT}
The attention mechanism in ViT plays a crucial role in capturing global dependencies and relationships among different regions of the input image. It enables the model to attend to relevant patches while processing the image sequence.

The core component of Transformer is the Self-Attention (SA) operation \cite{vaswani2023attention}, also known as "Scaled Dot-Product Attention". Assume that $X = [x_1, x_2, ...] \in \mathbb{R}^{N \times d}$ is an input sequence of $N$ elements/tokens. The sequence undergoes a preprocessing step, which includes positional encoding. The SA operation computes Query ($Q$), Key ($K$), and Value ($V$) embeddings from the input sequence using projection matrices:

\begin{equation}
Q = X W_Q, \quad K = X W_K, \quad V = X W_V,
\end{equation}

where $W_Q, W_K,$ and $W_V$ are projection matrices. The output of self-attention is computed as:

\begin{equation}
\text{SA}(Q, K, V) = \text{softmax}\left(\frac{QK^T}{\sqrt{d}}\right) V.
\end{equation}

In practice, multiple self-attention sub-layers are often stacked in parallel, forming a structure known as Multi-Head Self-Attention. Each head independently attends to different parts of the input sequence, and the outputs are concatenated and linearly transformed to obtain the final output.

To assist the decoder in learning contextual dependencies and to prevent attending to subsequent positions, a modification of self-attention known as Masked Self-Attention (MSA) is employed. It utilizes a masking matrix $M$ to restrict attention to previous positions:

\begin{equation}
\text{MSA}(Q, K, V) = \text{softmax}\left(\frac{QK^T}{\sqrt{d}} + M\right) V.
\end{equation}

The output of the attention sub-layers is then passed through a position-wise Feed-Forward Network (FFN), consisting of linear layers with non-linear activation functions such as ReLU or GELU.
The mathematical formulation of the ViT model can be expressed as:

\begin{equation}
\text{ViT}(X) = \text{FFN}(\text{MSA}(\text{SA}(X))),
\end{equation}

where $\text{SA}$ represents the Self-Attention mechanism, $\text{MSA}$ represents Masked Self-Attention, and $\text{FFN}$ denotes the Feed-Forward Network \cite{10123038, 9716741}.

We evaluated attention maps using the attention weights, $A_W$. These weights are computed from the model's outputs, including the $Q$ and $K$ tensors, through matrix multiplication followed by softmax normalization, as shown below:

\begin{equation}
A_W = \text{softmax}\left( Q \times K^{\top} \right) \times S
\end{equation}
This operation captures the importance of different patches in the input sequence. Scaling the resulting attention weights by a factor of $S$ enhances visualization of the attention patterns.

\subsection{Loss Functions}

As previously mentioned, contrastive methods can be employed in MAI. The objective is to learn an embedding space where similar sample pairs remain close to each other, while dissimilar ones are far apart.
The contrastive loss function \cite{1467314} is one of the initial training objectives employed in similarity metric learning. Given a list of input samples $x_i$, each with a corresponding label $y_i$ among $C$ classes, we aim to learn a function $f$ that encodes $x_i$ into an embedding vector. This function should ensure that examples from the same class have similar embeddings, while samples from different classes have very different embeddings. Consequently, the contrastive loss takes a pair of inputs $(x_i, x_j)$ and minimizes the embedding distance when they are from the same class but maximizes the distance otherwise.

\begin{equation}
\label{eq:con}
\begin{split}
\mathcal{L}_{\text{con}}(x_i, x_j, y_i, y_j) = & \frac{1}{2} (1 - y_{ij}) d^2 \\
& + \frac{1}{2} y_{ij} \max(0, m - d^2)
\end{split}
\end{equation}
where $d$ is the Euclidean distance between the embeddings of $x_i$ and $x_j$, and $m$ is a hyperparameter defining the lower bound distance between samples of different classes.

When label data is available, Supervised Contrastive Loss \cite{khosla2021supervised} is more efficient then SSL approach by imposing normalized embeddings from the same class to be closer together than embeddings from different classes. Given a set of randomly sampled $n$ (image, label) pairs,  $\{x_i, y_i \}_{i=1}^n$, $2n$ training pairs can be created by applying two random augmentations of ever sample, $\{\tilde{x}_i, \tilde{y}_i\}_{i=1}^n$. Similar to Soft Nearest Neighbor Loss \cite{frosst2019analyzing}, $\mathcal{L}_{\text{supcon}}$ uses multiple positive and negative samples:

%\begin{equation}
%\begin{split}
%\label{eq:supcon}
%\mathcal{L}_{\text{supcon}} = & - \sum_{i=1}^{2n} \frac{1}{2|N_i|-1} \sum_{j \in N(y_i), j \neq y_i} \\
%& \log \frac{\exp(\mathbf{z}_i \cdot \mathbf{z}_j / \tau)}{\sum_{k \in I, k\neq i} \exp(\mathbf{z}_i \cdot \mathbf{z}_k / \tau)} 
%\end{split}
%\end{equation}

\begin{subequations}
\label{eq:contrastive_loss}
\begin{align}
\mathcal{L}_{\text{supcon}} &= - \sum_{i=1}^{2n} \frac{1}{2|N_i|-1}  \mathcal{L}_{\text{supcon}}^i \\
\mathcal{L}_{\text{supcon}}^i &= \sum_{\substack{j \in N(y_i), \\ j \neq y_i}} \log \frac{\exp(\mathbf{z}_i \cdot \mathbf{z}_j / \tau)}{\sum_{\substack{k \in I, \\ k\neq i}} \exp(\mathbf{z}_i \cdot \mathbf{z}_k / \tau)}
\end{align}
\end{subequations}
where $\mathbf{z}_k = P(E(\tilde{x}_k))$, in which $E(.)$ is an encoder network and $P(.)$ is a projection network. The set $N_i = { j \in I : \tilde{y}_j = \tilde{y}_i }$ consists of indices corresponding to samples in $I$ with the label $\tilde{y}_i$. Enlarging the set $N_i$ by including more positive samples enhances the performance of the model.

% ranked SupRL
Based on $\mathcal{L}_{\text{supcon}}$, Zha et al. \cite{NEURIPS2023_39e9c591} introduced the Rank-N-Contrast Loss ($\mathcal{L}_{\text{RnC}}$) for regression-aware representation learning of continuous and ordered samples. To align distances in the embedding space with the ordering of distances in their labels $\mathcal{L}_{\text{RnC}}$ ranks samples based on their target distances and contrasts them against each other using their relative rankings. 
Given an anchor $v_i$, $\mathcal{L}_{\text{RnC}}$ models the likelihood of any other $v_j$ to increase exponentially with respect to their similarity in the representation space. $\mathcal{L}_{\text{RnC}}$ introduces $S_{i,j} := \{ v_k | k = i, d(\tilde{y}_i, \tilde{y}_k) \geq d(\tilde{y}_i, \tilde{y}_j) \}$ to denote the set of samples that are of higher ranks than $v_j$ in terms of label distance with respect to $v_i$, where $d(\cdot,\cdot)$ is the distance measure between two labels (e.g., $L_1$ distance). Then the normalized likelihood of $v_j$ given $v_i$ and $S_{i,j}$ can be written as:

\begin{equation}
\label{eq:rnc1}
\mathbb{P} (v_j | v_i, S_{i,j}) = \frac{\exp(\text{sim}(v_i, v_j)/\tau)}{\sum_{v_k \in S_{i,j}} \exp(\text{sim}(v_i, v_k)/\tau)} 
\end{equation}
where $\text{sim}(\cdot,\cdot)$ represents the similarity measure between two feature embeddings (e.g., negative $L_2$ norm), and $\tau$ denotes the temperature parameter. For a given random batch, and for a given anchor, we select the positive pairs based on the minimum $L_1$ distance and assign negative pairs if the distance is greater.

In this work, instead of using augmentations, we utilize data from multiple views of the same object (melt pool) in a multimodal fashion. As we have two process parameters, we perform contrastive losses for each in a separate embedding. By incorporating different modalities, we also minimize the risk of model collapse.

%\begin{equation}
%\label{eq:supcon}
%\mathcal{L}_{\text{supcon}} = \frac{1}{2n} \sum_{i=1}^{2n} \log \frac{\exp(\mathbf{Z}_i \cdot \mathbf{Z}_j / \tau)}{\sum_{k \in E_k(i)} \exp(\mathbf{Z}_i \cdot \mathbf{Z}_k / \tau)}
%\end{equation}

\section{Proposed Method}

We present a new approach (JEMA) that leverages metadata such as process parameters to create a representation that embeds this process knowledge. This multimodal co-learning framework was developed to simultaneously train the representation and the final task. We intend to use this representation to make model predictions robust and explainable using only the main modality: on-axis data.

\subsection{JEMA Architecture}

\begin{figure}[h!]%
\centering
\includegraphics[width=0.4\textwidth]{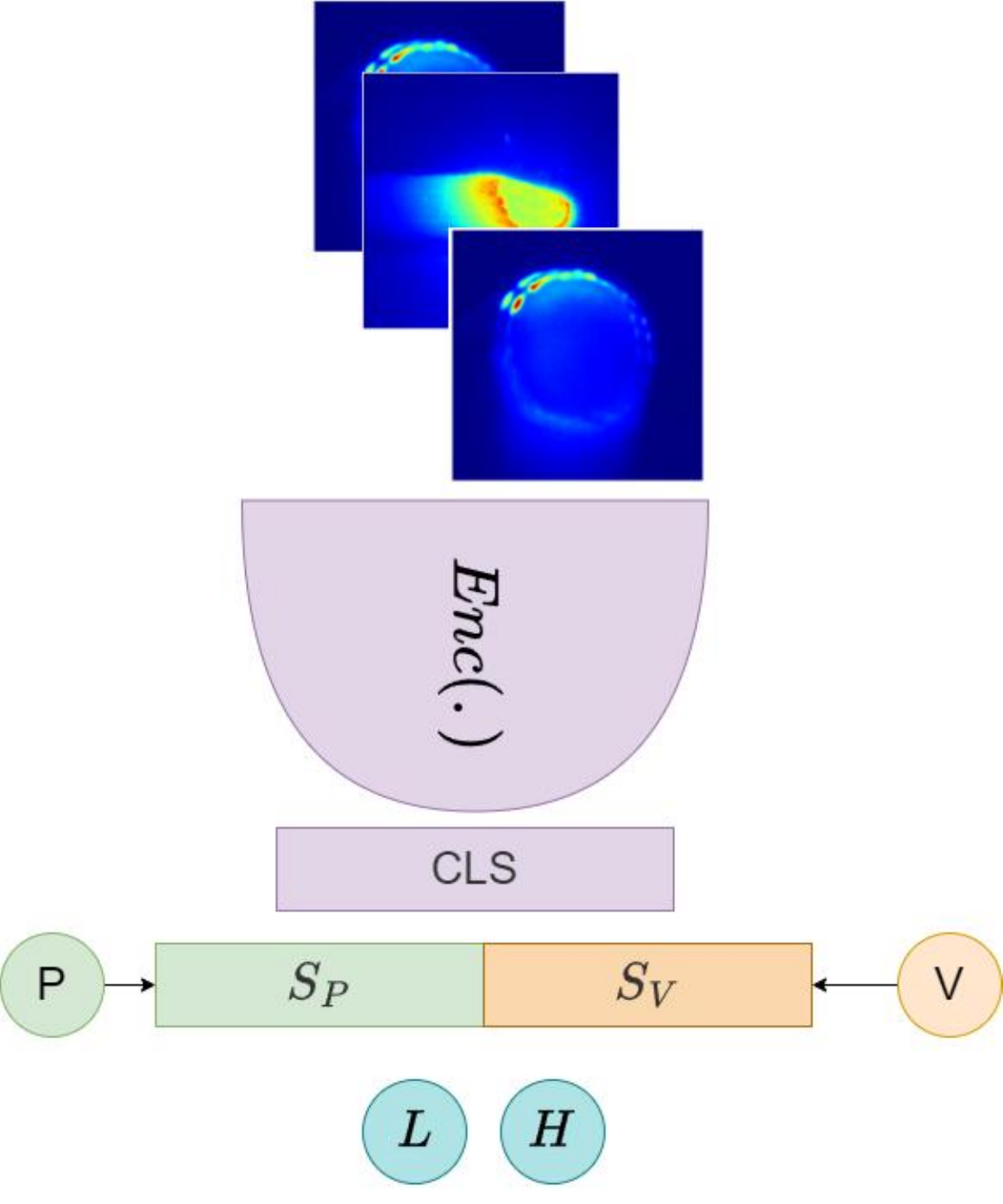}
\caption{Diagram of JEMA architecture showing different modalities as input and the embedding representation with the process parameters such as laser power (P) and velocity (V). The model predicts the MP Length (L) and Height (H) as a quality measure of the process.}\label{jema_1}
\end{figure}

Our problem is characterized by having a multimodal image dataset with metadata. We use the vanilla ViT model proposed by Dosovitskiy et al. \cite{dosovitskiy2021image}, based on the base-size architecture with a patch resolution of 16x16 for image processing. To feed images to the Transformer encoder, each image is split into a sequence of fixed-size non-overlapping patches, which are then linearly embedded. A Classify Token (CLS) is added to represent an entire image, which is used to predict the embedding. The embedding layer appended to the output of the CLS token in the ViT serves as a predictive mechanism. It consists of two layers, each containing 128 neurons, used to embed metadata such as laser power and velocity: $S_P$ and $S_V$ as shown in Figure~\ref{jema_1}. Subsequently, from the combined embedding, a linear layer is added to predict the final task, specifically the melt pool dimensions including L and H.

%%% IMAGEM %%%%%

%"A Survey on Vision Transformer"
%"Transformers in Vision: A Survey"

\subsection{JEMA Loss}

The JEMA architecture facilitated the development of a loss function that combines regression loss for the final prediction with supervised contrastive loss for the embeddings $S_P$ and $S_V$. JEMA Loss is a supervised similarity loss that is regression-aware and relies directly on the distance between the metadata, instead of using ranking algorithms. In this view, $\mathcal{L}_{\text{CR}}$ is the regression loss between the cosine similarity of embeddings and the square difference between embeddings. We formulate the $\mathcal{L}_{\text{CR}}$ as: given embedding vector $\mathbf{x}_i$ and corresponding metadata $u_i$ where $i = j = 1, 2, ..., N$, the pair wise similarity matrix is given by the un-normalize cosine similarity:

\begin{equation}
    \text{sim}(\hat{\mathbf{x}}_j, \hat{\mathbf{x}}_i) = 1 - x_i \cdot z_j^T
\end{equation}
Other similarity metrics, $\text{sim}(\cdot,\cdot)$, such as L2- or L1-Distance are also evaluated. Then, the pairwise squared differences between labels are computed as:

\begin{equation}
\text{D}(i, j) = (u_i - u_j)^2
\end{equation}
The Mean Squared Error (MSE) loss between the similarity matrix and the labels difference matrix is computed as:

\begin{equation}
\mathcal{L}_{CR} = \frac{1}{N} \sum_{i=1}^{N} \sum_{\substack{j=1 \\ j \neq i}}^{N}  \Bigr[ \text{sim}(\cdot,\cdot) - \text{D}(i, j) \Bigr] ^2
\end{equation}
This loss function penalizes the model when the pairwise similarity of feature vectors deviates from the squared differences of their corresponding labels.The diagonal entries of both the similarity matrix, denoted $\text{sim}(\cdot,\cdot)$, and the distance matrix, denoted $\text{D}(i, j)$, were excluded due to their trivial nature, where the similarity or distance between an item and itself is inherently zero. The goal is to minimize this loss, encouraging the model to learn feature representations that preserve the relationships between the input data and their labels.

The separate regression loss for melt pool L and H is computed using standard regression loss functions, such as MSE, and is denoted as $\mathcal{L}_{\text{reg}}^L$ and $\mathcal{L}_{\text{reg}}^H$, respectively. Considering the a label $y$ and corresponding prediction $\hat{y}$, the $\mathcal{L}_{\text{reg}}$ is

\begin{equation}
\mathcal{L}_{\text{reg}} = \frac{1}{N} \sum_{i=1}^{N} \Bigr[ y - \hat{y} \Bigr] ^2
\end{equation}

The overall loss function, denoted as $\mathcal{L}_{\text{JEMA}}$, is a weighted combination of the contrastive regression losses and the regression losses for melt pool L and H:

\begin{equation}
\mathcal{L}_{\text{JEMA}} = \alpha \left( \mathcal{L}_{\text{CR}}^P + \mathcal{L}_{\text{CR}}^V \right) + \beta \left( \mathcal{L}_{\text{reg}}^L + \mathcal{L}_{\text{reg}}^H \right)
\end{equation}

where $\alpha$ and $\beta$ are hyperparameters controlling the weights of the contrastive and regression loss terms, respectively.

\section{Experiment and Results}

This section explains the experimentation details and presents the main results, including regression metrics and embedding representations.

\subsection{Implementation Details}

The JEMA architecture was trained using various loss functions (SupCon, RnC, and JEMA) and compared with a simple regression approach. Data augmentation was implemented, incorporating the following random transformations: resizing (from 320x320 to 224x224 image size), horizontal and vertical flipping, random rotation ($\pm$20º), and shifting (5\%). The model's training was performed on a GPU cluster equipped with a Nvidia A6000 and 96 GB of RAM. The PyTorch library was employed for the development and training of these models. It is essential to highlight that in all approaches, we deliberately chose not to normalize the embeddings. This decision was made as normalizing them adversely affected performance on our dataset.

\subsection{Co-Learning Results}

In Table~\ref{tab:1}, we present  MSE results for predicting melt pool L and H, alongside a comparison to the performance of the multimodal regression approach. The data clearly illustrate that employing a contrastive or regularisation approach to align the multimodal data instead of simple regression yields significant improvements. Notably, the JEMA approach showcases remarkable enhancements within the multimodal context, with its benefits even extending to the unimodal setting, exhibiting an improvement from -5\% to -4\% compared to the SupCon loss.

\begin{table}[htbp]
  \centering
  \begin{center}
  \renewcommand{\arraystretch}{1.5} % Adjust the value as needed
  \caption{Comparison of MSE and its variation on different loss functions.}
    \begin{tabular}{cccccc}
    \hline
          &       & \multicolumn{2}{c}{MSE ($\times 10^{-4}$)} & \multicolumn{2}{c}{Variation (\%)} \\
    \cline{3-4} \cline{5-6}
          &       & Multi- & Uni- & Multi- & Uni- \\ 
    \hline
           & Reg   & $3.27$ & $4.22$ & 0 & -29 \\
    Others & SupCon & $2.57$ & $3.44$ & 21 & -5 \\
          & RnC & $2.60$ & $3.93$ & 20 & -20 \\ 
    \hline
    Ours   & Cosine & $2.32$ & $3.39$ & \textbf{29}  & \textbf{-4} \\
    (JEMA)             & L2-distance & $2.53$ & $3.63$ & 23  & -11 \\
                 & L1-distance & $3.87$ & $4.85$ & -18 & -48  \\ 
    \hline
    \end{tabular}%
  \label{tab:1}%
  \end{center}
\end{table}%

\begin{figure}[h!]
\centering

\subfloat[]{\includegraphics[width=0.35\textwidth]{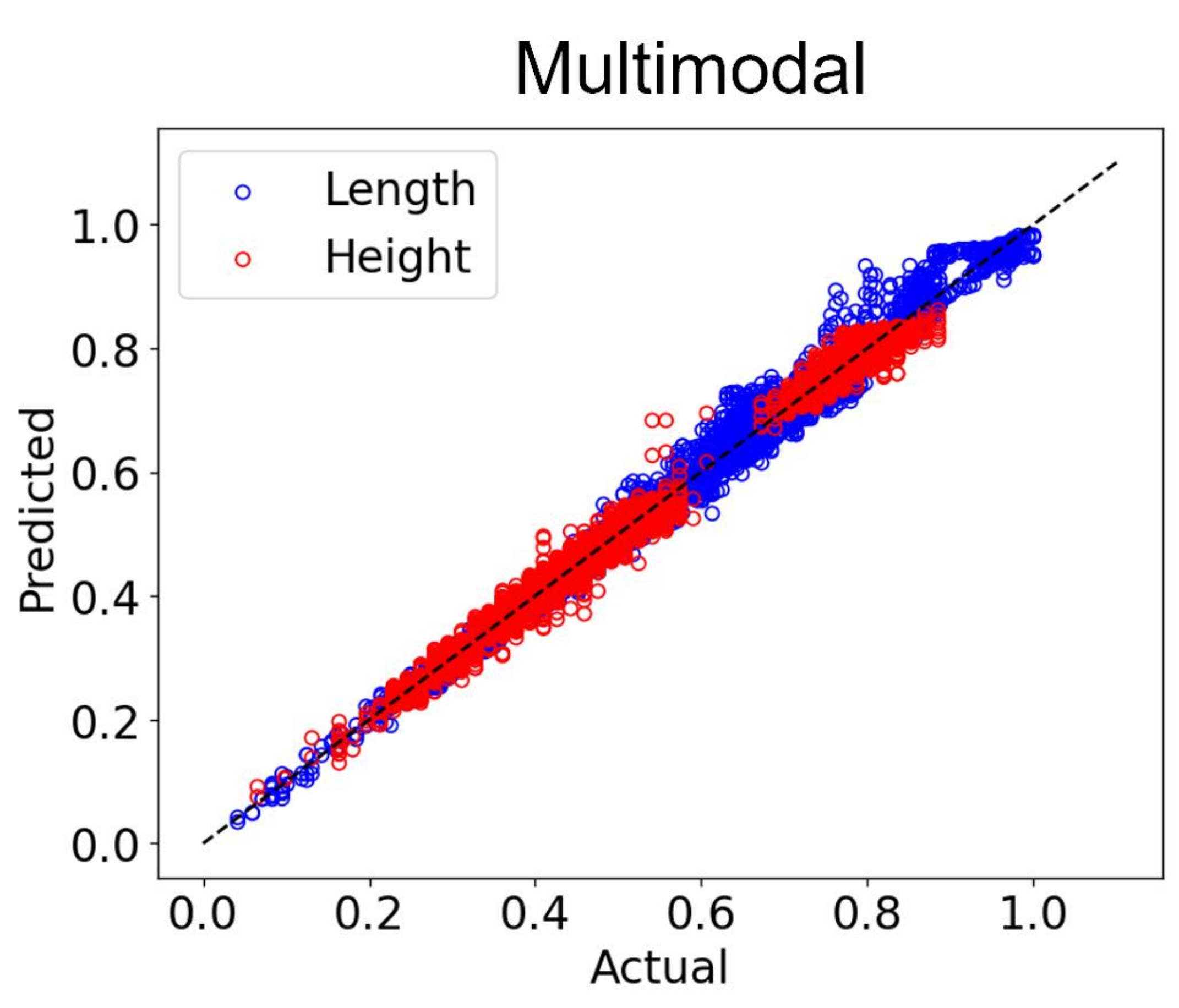}}
\hfil
\subfloat[]{\includegraphics[width=0.35\textwidth]{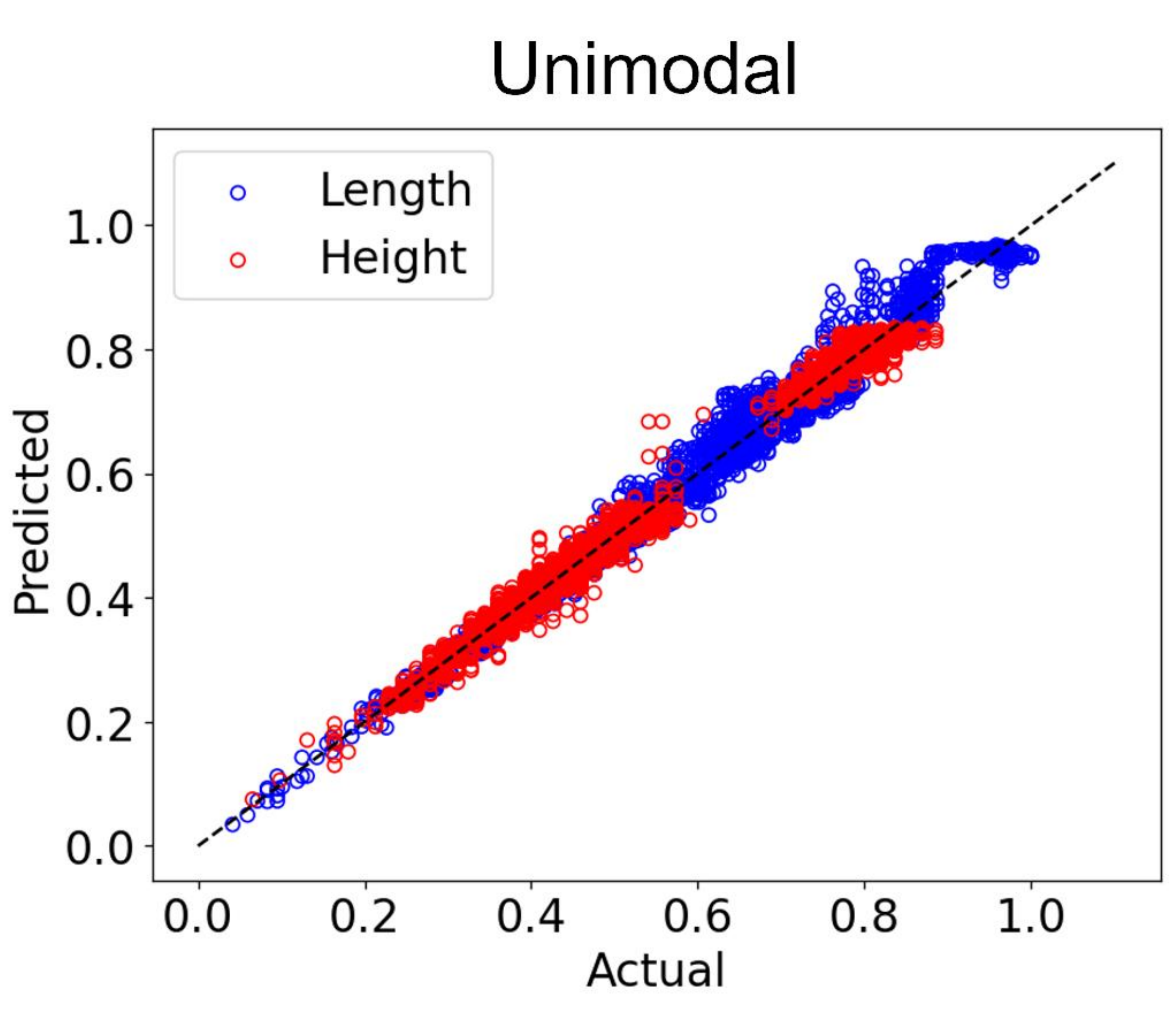}}
\caption{Comparison of the melt pool length and height predictions with the JEMA cosine loss function. (a) Multi-modal setting, and (b) Unimodal setting.}
\label{jema_pred}
\end{figure}

In Figure~\ref{jema_pred}, we present plots illustrating the predicted and actual values of melt pool L and H using our approach, in both multimodal and unimodal settings. Additionally, plots for each approach can be found in Appendix~\ref{ApxA}. Notably, in the JEMA approach, fewer outliers are observed compared to the SupCon method.

%%% IMAGEM %%%%%

\subsection{Embedding Representation}

\begin{figure*}[h!]
\centering
\includegraphics[width=1.0\textwidth]{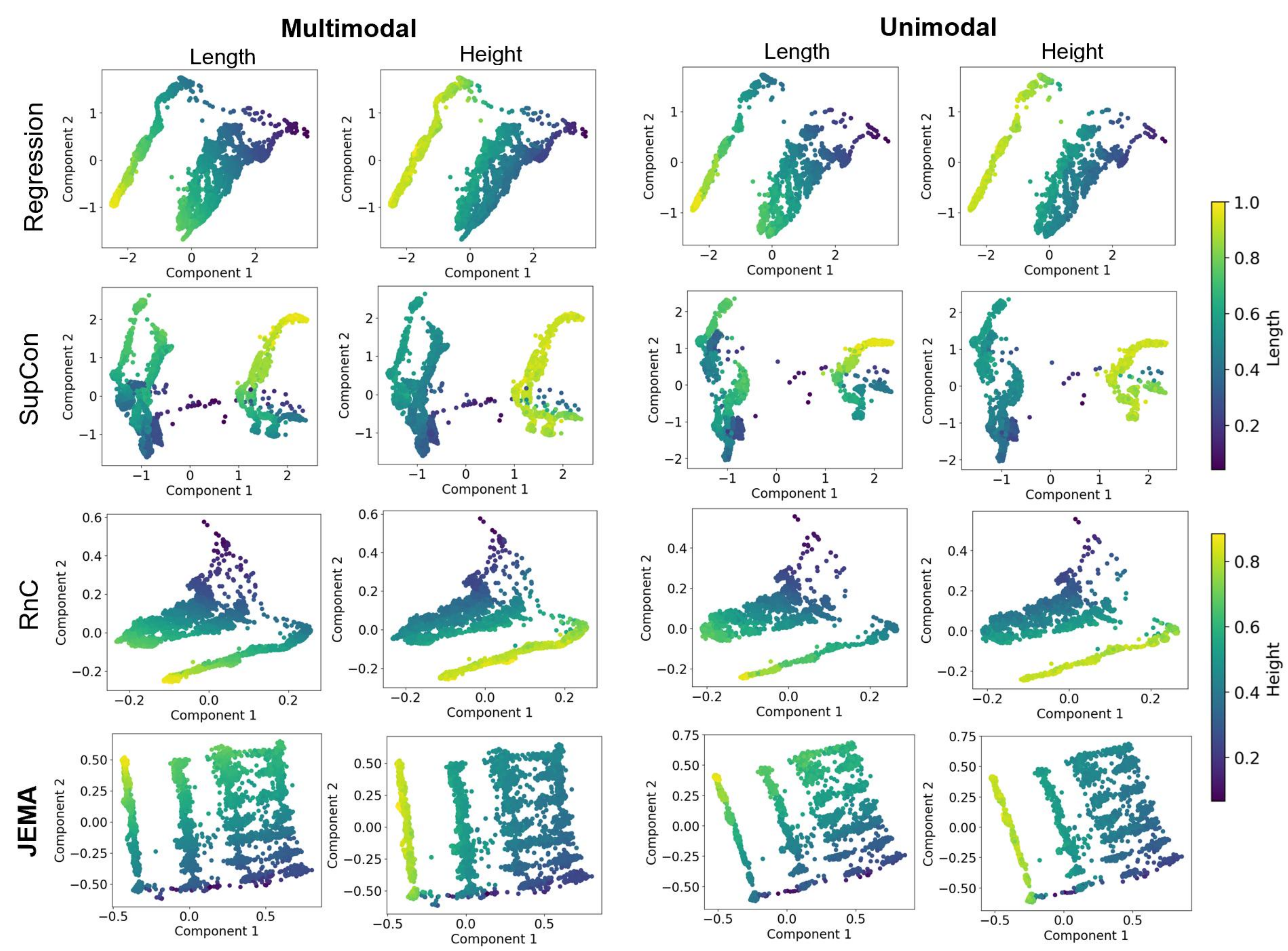}
\caption{Visual representation of embedding using Principal Component Analysis (PCA) with 2 components.}
\label{pca}
\end{figure*}

The Principal Component Analysis (PCA) and t-Distributed Stochastic Neighbor Embedding (t-SNE) techniques were employed to analyze the embedding representations, each with 2 components. Figure~\ref{pca} shows the PCA representation for each approach. Our approach has simpler and distributed shapes which allows us to evaluate them more easily. The t-SNE representations are included in the Appendix~\ref{ApxB}. 
%[The color represents the predictions (melt pool L and H).] 
%%% IMAGEM %%%%%
%[Color indicates predictions for melt pool length and height.] - ADD color bar}

\begin{figure}[h!]
\centering

\subfloat[]{\includegraphics[width=0.4\textwidth]{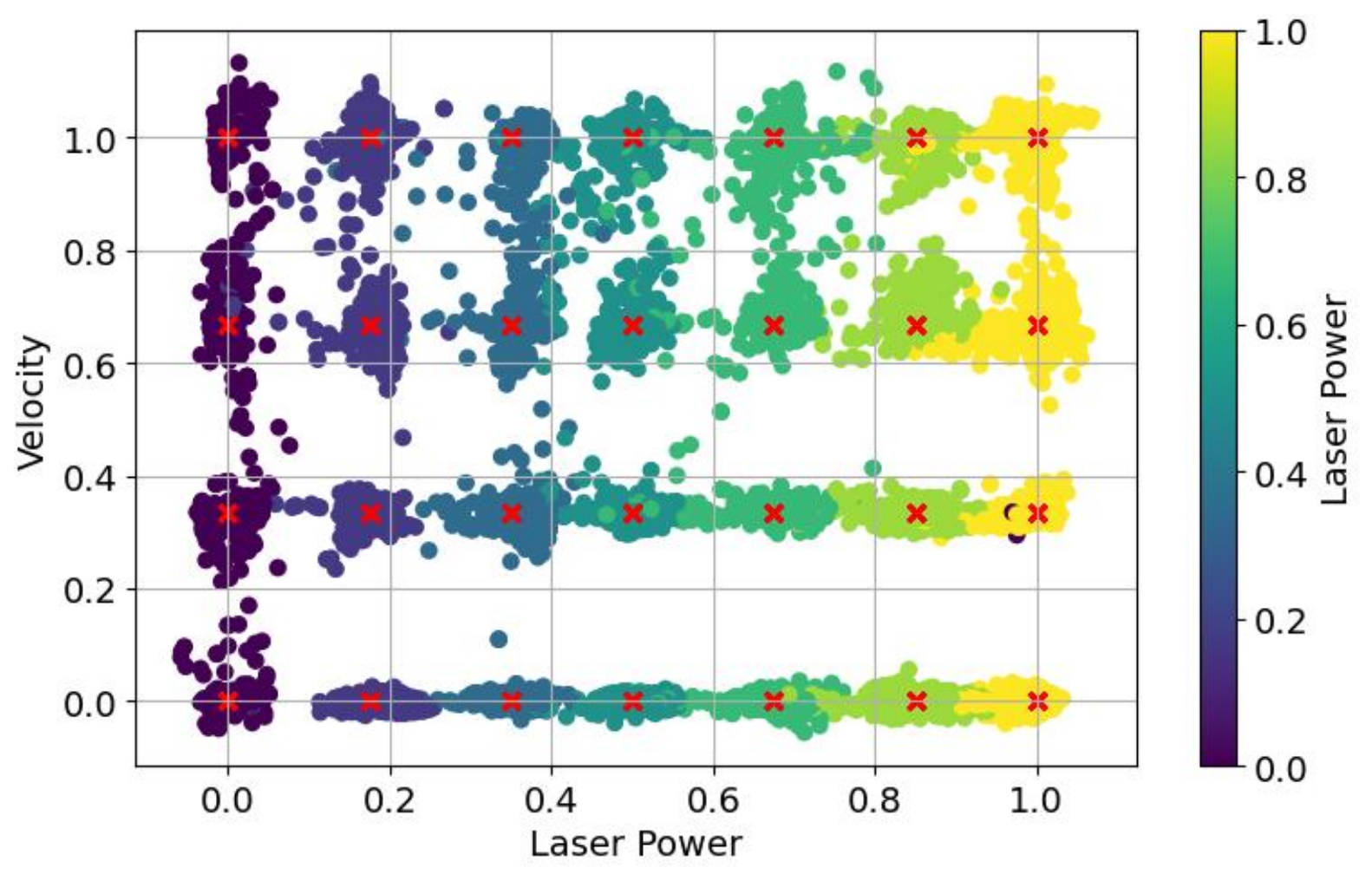}}
\hfil
\subfloat[]{\includegraphics[width=0.4\textwidth]{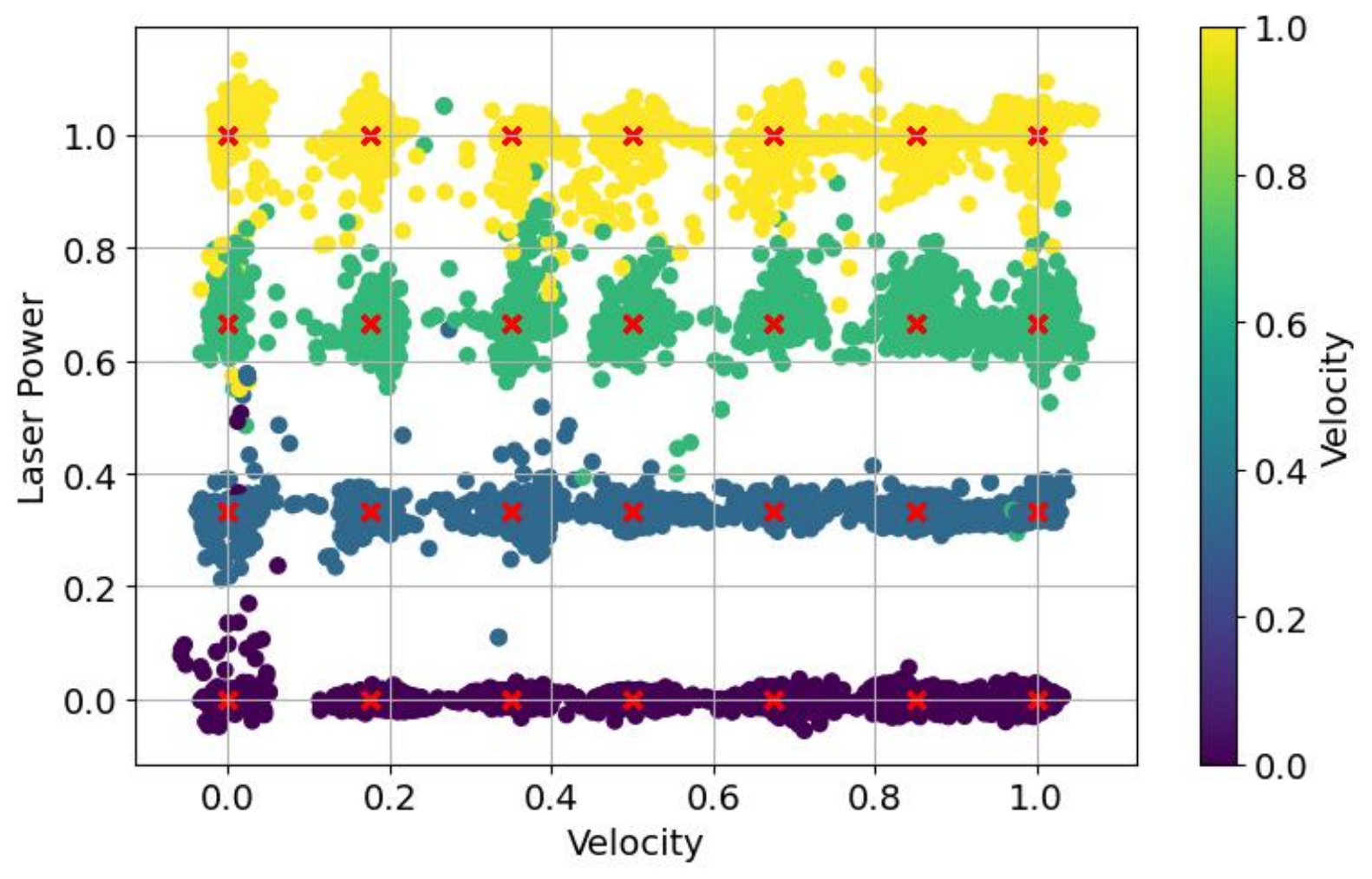}}

\vspace{0.5em} % Adjusts vertical spacing between rows

\subfloat[]{\includegraphics[width=0.4\textwidth]{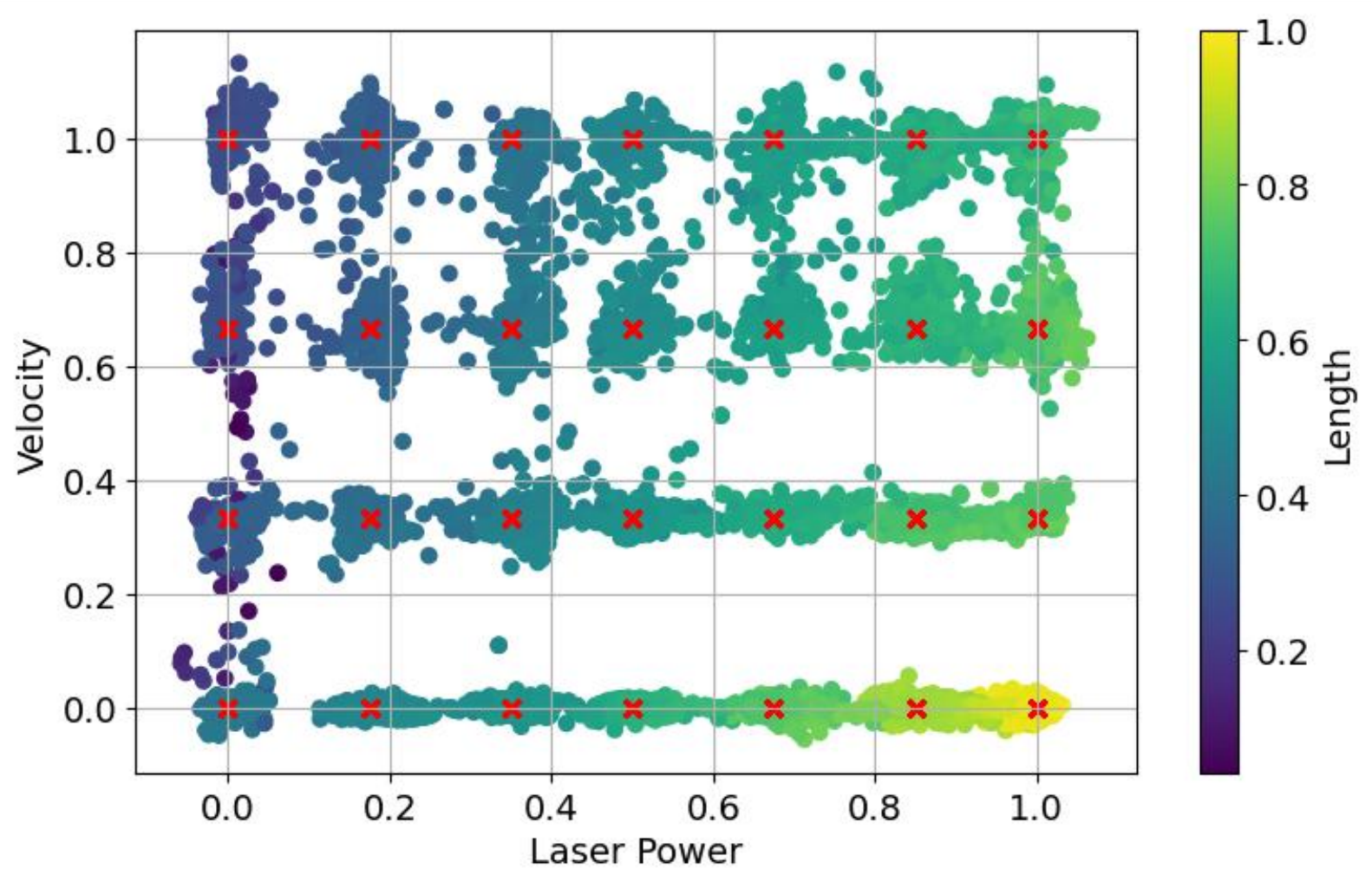}}
\hfil
\subfloat[]{\includegraphics[width=0.4\textwidth]{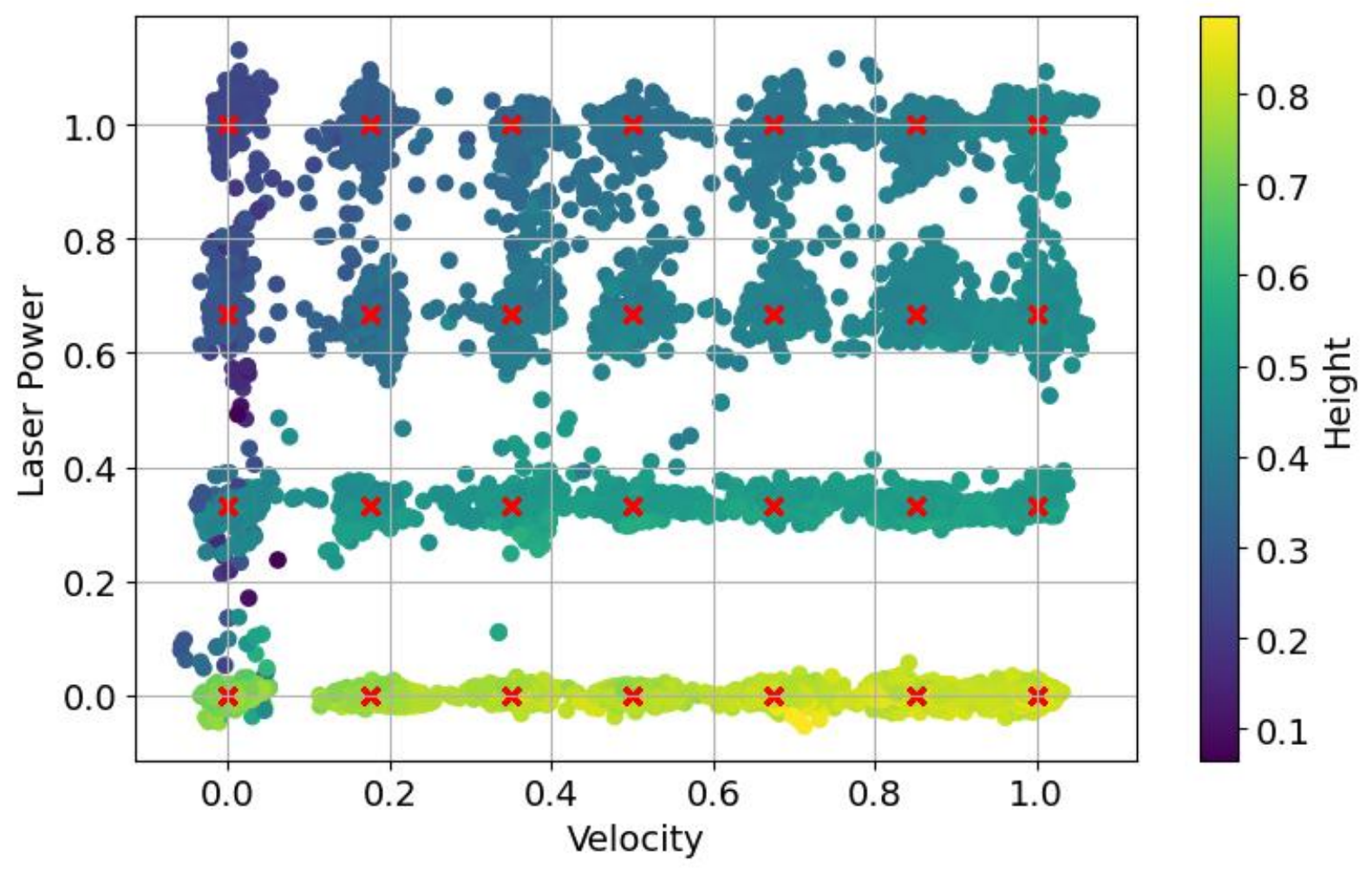}}

\caption{Evaluation of JEMA's embedding representation in the primary modality (on-axis) using linear regression. In (a), the representation of laser power is assessed; (b) shows the representation of velocity; (c) displays the predictions for melt pool length; and (d) illustrates the predictions for height.}
\label{fig_jema_1}
\end{figure}

For a deeper understanding of our JEMA representations, we utilized linear regression with scikit-learn \cite{scikit-learn} to predict the process parameters (\(u\)). As illustrated in Figure~\ref{fig_jema_1}, each circular point represents an embedding derived from an image, while the red 'x' marks denote the true parameters, which serve as our targets. It is observed that both modalities formed a similar embedding representation, indicating the effectiveness of our approach.
This visualization also demonstrates a successful direct inference of the tested process parameters based on the embeddings. Furthermore, as shown in Figures~\ref{fig_doe} and~\ref{pca}, the two-component PCA primarily captures the characteristics of velocity and laser power.
We also observe that lower velocities are more readily distinguished, as the melt pool shape tends to vary less at higher velocities, leading to a denser concentration of data points at lower velocity ranges. Additionally, we have more data for these process parameters compared to the high velocities. More data needs to be added to assess if it could further enhance the analysis.

\begin{figure}[h!]
\centering

\subfloat[]{\includegraphics[width=0.35\textwidth]{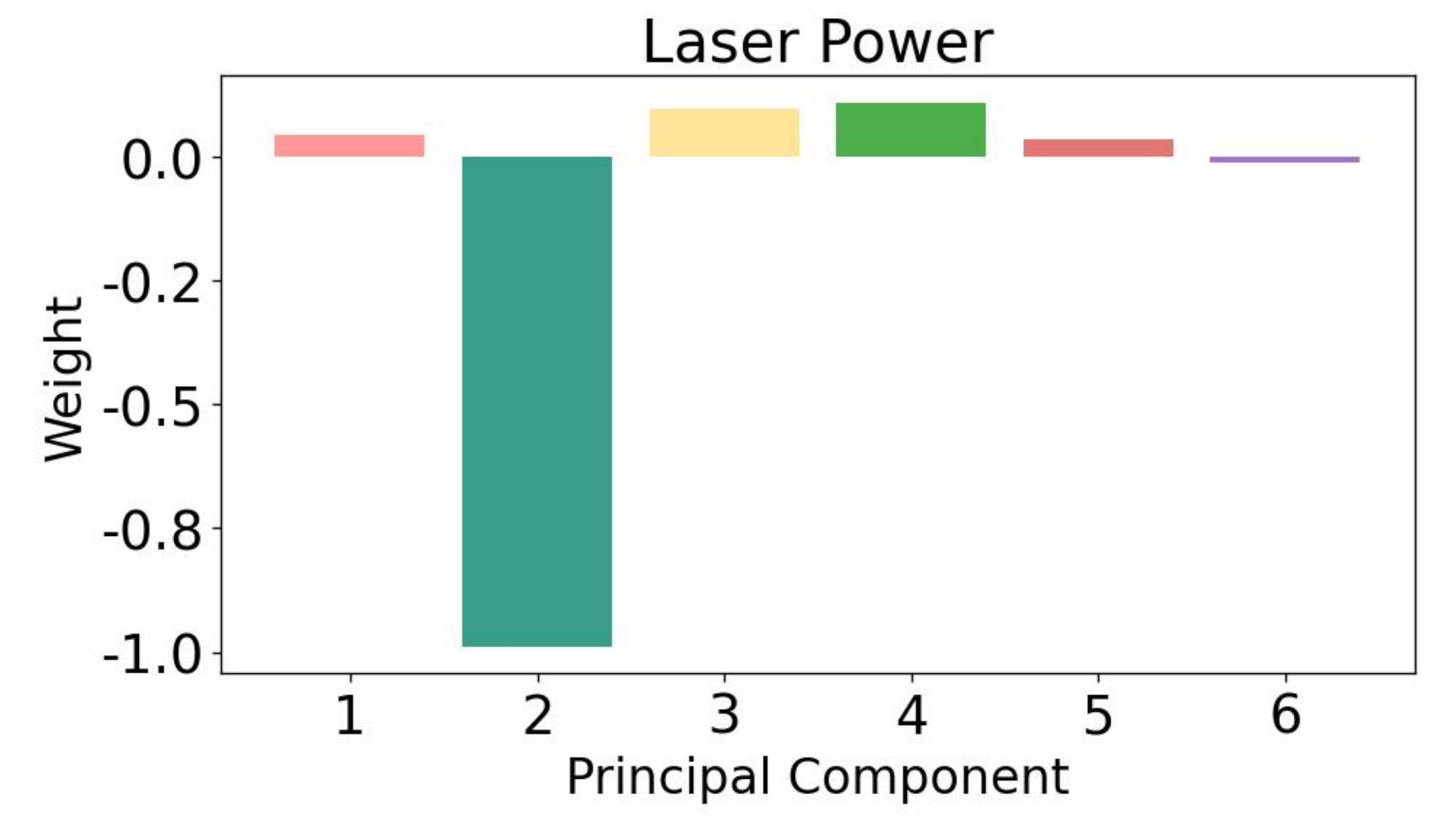}}
\hfil
\subfloat[]{\includegraphics[width=0.35\textwidth]{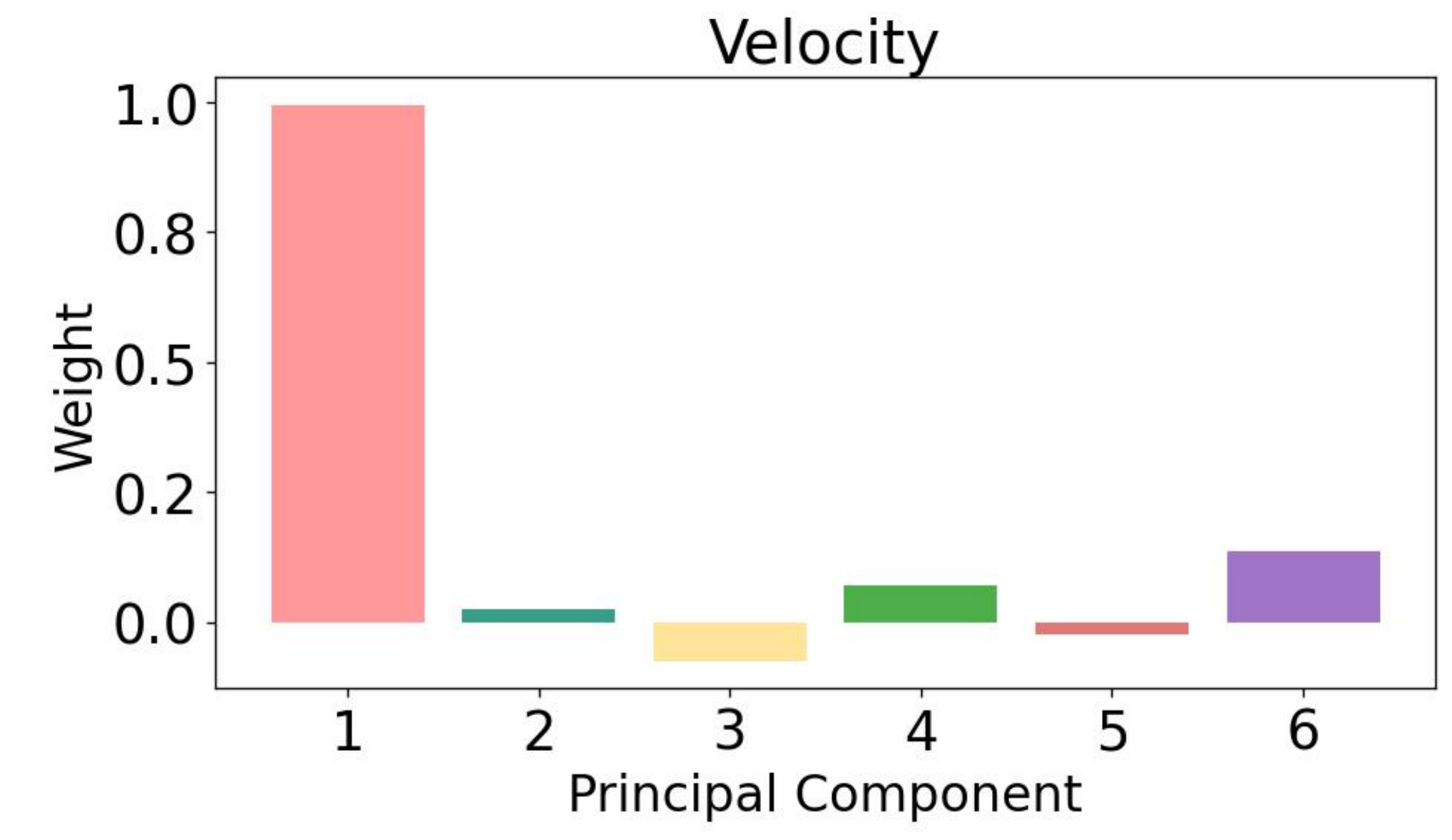}}

\vspace{0.5em} % Adjusts vertical spacing between rows

\subfloat[]{\includegraphics[width=0.35\textwidth]{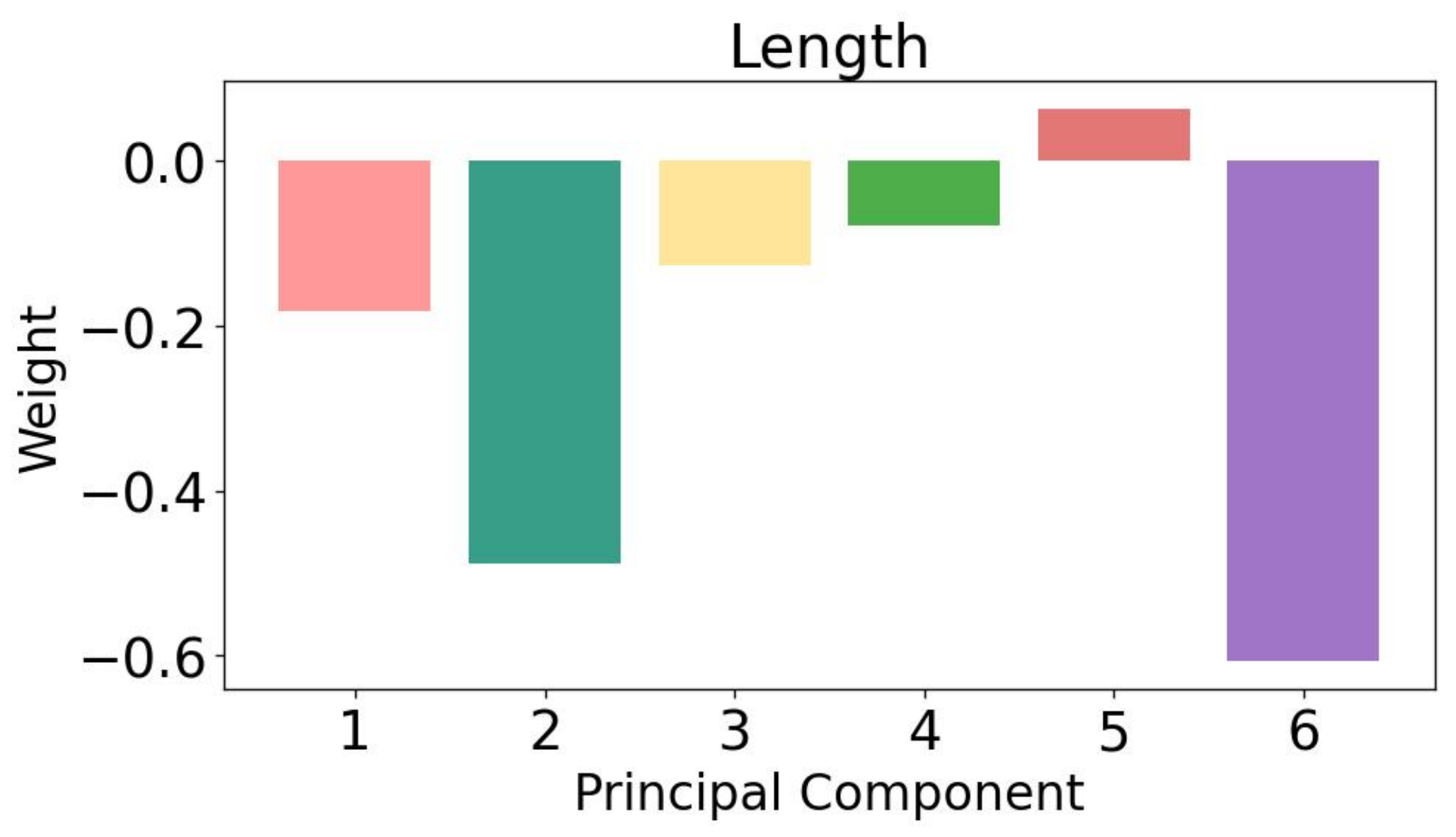}}
\hfil
\subfloat[]{\includegraphics[width=0.35\textwidth]{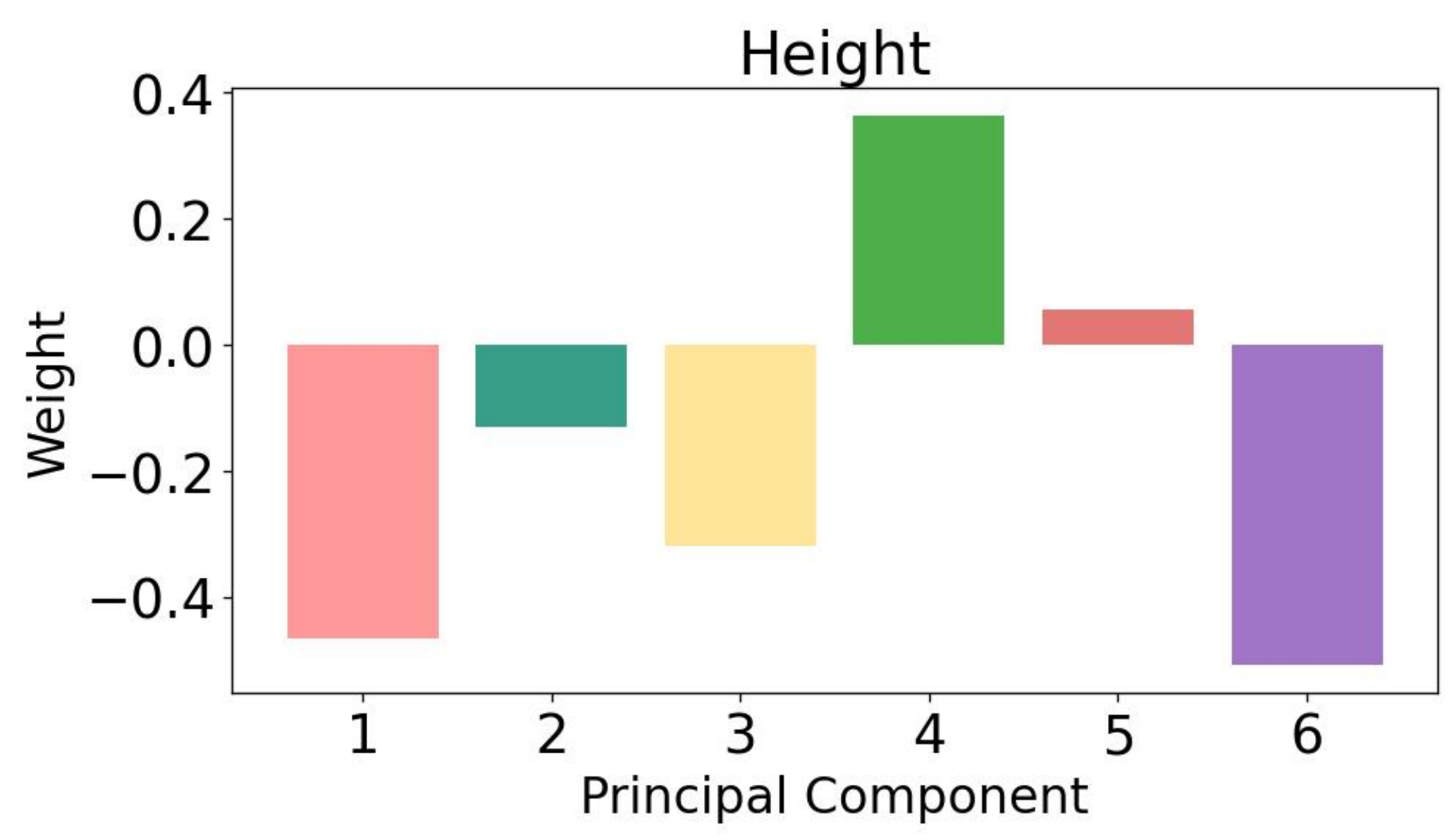}}

\caption{Evaluation of the importance of each component in predicting (a) laser power, (b) velocity, (c) melt pool length, and (d) height.}
\label{PCA+linear}
\end{figure}

Furthermore, to evaluate the significance of the PCA components, we conducted PCA fitting to achieve 95\% accuracy and obtained six components. By employing linear regression to predict the process parameters, melt pool length (L) and height (H), we can assess the importance of each component, as depicted in Figure~\ref{PCA+linear}. It reveals that the first component primarily represents velocity, while the second component corresponds to laser power. This evaluation is promising not only from a theoretical standpoint but also bears practical significance in real-world manufacturing scenarios. Understanding the significant relationship between melt pool length and laser power, as well as melt pool height and velocity, enables us to validate existing literature and deepen our comprehension of the manufacturing process. Furthermore, this insight lays the groundwork for optimizing manufacturing processes and potentially implementing control strategies to enhance efficiency and process quality.

%Additionally, we observe a strong dependency of melt pool length on laser power and melt pool height on velocity. This evaluation enabled us to understand the relationship between process parameters and the predictions.

%%% IMAGEM %%%%%
%\begin{figure}[!]
%\centering
%\includegraphics[width=0.7\textwidth]{jema_3.pdf}
%\caption{Evaluation of the importance of each component in predicting laser Power, velocity, melt pool length, and height.}
%\label{PCA+linear}
%\end{figure}%

\subsection{Attention Maps}

\begin{figure}[!]
\centering
\includegraphics[width=0.55\textwidth]{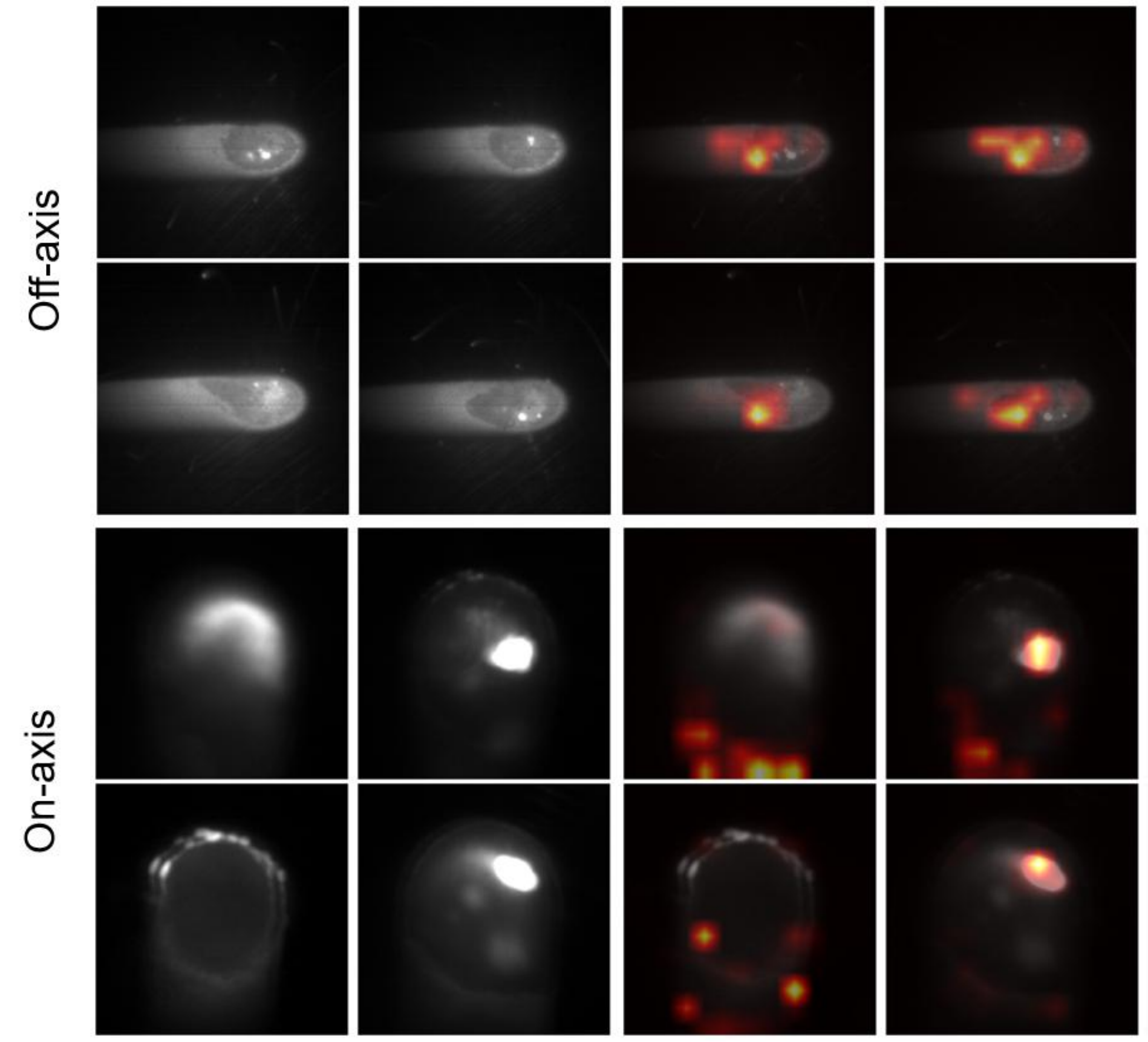}
\caption{Attention maps of the last layer from the ViT JEMA model. On the left is the original image, while on the right is the overlapped attention map, illustrating the model's focus areas.}
\label{att_maps_1}
\end{figure}

In Section III, we discussed how ViTs utilize attention mechanisms to understand which parts of an image are most important. This helps the model focus on the key areas while processing the image. Figure~\ref{att_maps_1} displays attention maps alongside original images from both off-axis and on-axis cameras. In the off-axis view, the model concentrates on the melt pool region, whereas in the on-axis view, it focuses on the edges and tail of the melt pool, as well as reflections of the liquid metal. This indicates that the model effectively learns to identify and prioritize relevant features in the images. In Figures~\ref{attmap_1a} and \ref{attmap_1b} in Appendix~\ref{ApxC}, the complete batch of on-axis and off-axis images is shown.

%%% IMAGEM %%%%%

In addition to its ability to identify relevant features, the model's capability to focus on specific regions bears significant implications for various applications. For instance, in industrial settings such as metal AM, understanding the dynamics of the melt pool is crucial for ensuring product quality and process optimization. By accurately capturing features like the melt pool region and edges, the model demonstrates its potential to assist in real-time process monitoring. Moreover, its ability to adapt its attention based on the camera perspective—such as off-axis and on-axis views—underscores its versatility and robustness across different imaging conditions. This not only enhances the model's performance but also broadens its applicability across diverse domains beyond manufacturing, such as medical imaging, autonomous vehicles, and remote sensing.

\section{Discussion}

In the domain of Industry 5.0, process monitoring plays a crucial role by providing real-time insights. Specifically, melt pool monitoring offers valuable information about process quality, enhancing our understanding of the manufacturing process. However, laser additive manufacturing, in particularly LMD, is intricate due to its involvement with multi-physics phenomena, and relying on a single sensor may limit our perception of the process dynamics. While data fusion has been utilized in various studies, its integration with ML models often overlooks interpretability.
Our proposed approach addresses this limitation by integrating and inferring meta-data, such as process parameters. This is particularly significant since the conditions during printing coats and parts may vary, rendering initial parameter optimization unreliable. Consequently, our method enables the prediction of melt pool features and the inference of parameterized process parameters.

Although our method shows promising results, it has only been tested with data from single track lines and a single material. Future research will evaluate its performance with different materials and the inclusion of real and simulated data. Additionally, exploring alternative ViT or hybrid models, as well as optimizing the embedding space dimension, are avenues for further investigation.

We anticipate that our model could also be applied for process control to enhance melt pool stability. However, we acknowledge the need to measure model inference time and optimize computational resources, especially considering the potential computational expense of ViT models.
While monitoring with cameras is common in LMD, the integration of other sensors such as pyrometers and audio sensors remains unexplored. Furthermore, exploring strategies for combining different data types could improve overall monitoring effectiveness.

Although successful in data representation, data transfer efficiency could be further enhanced. Increasing the embedding size may improve this aspect but must be balanced with considerations of model complexity and the risk of overfitting. These considerations are important for the practical implementation of our approach in industrial settings.

\section{Conclusion}
In this study, we introduced JEMA, a framework for multimodal co-learning designed to align the embedding space using meta-data such as process parameters. This alignment facilitates the inference of similarity to compare with the tested process parameters, enabling direct comparison with initial process parameters and enhancing our understanding of the manufacturing process.
Our experimentation in the LMD process demonstrated that JEMA surpasses existing methods in both multi and unimodal settings. The acquired data representation not only enables inference of metadata but also offers insights into initial process parameters.

Furthermore, the flexibility of JEMA allows for potential adaptation and extension beyond the LMD process. Its modular architecture enables seamless integration with various manufacturing processes, thus extending its applicability beyond additive manufacturing. By harnessing expertise from researchers and practitioners across different domains, JEMA holds the potential to drive advancements in process monitoring and optimization, not only in Industry 5.0 but also in other fields.
Although further testing of JEMA is warranted, we encourage the community to explore and implement this framework in diverse environments.

\section*{Acknowledgments}
Open access funding provided by FCT-FCCN (b-on). 

The authors acknowledge João Sousa financial support from Fundação para a Ciência e a Tecnologia (FCT), Portugal, under Grant number 2022.09967.BD. Additionally, the authors gratefully acknowledge the funding of Project Hi-rEV - Recuperação do Setor de Componentes Automóveis (C644864375-00000002) cofinanced by Plano de Recuperação e Resiliência (PRR), República Portuguesa, through NextGeneration EU.

The authors gratefully acknowledge the support of the Institute of Science and Innovation in Mechanical and Industrial Engineering (INEGI) for providing the machine setup and computational resources essential to this work.

%Bibliography
\bibliographystyle{unsrt}  
\bibliography{references}

% Clear the page and start the appendix
\clearpage

\appendix

\section{Appendix A}
\label{ApxA}

\begin{figure*}[h!]
\centering
\includegraphics[width=0.60\textwidth]{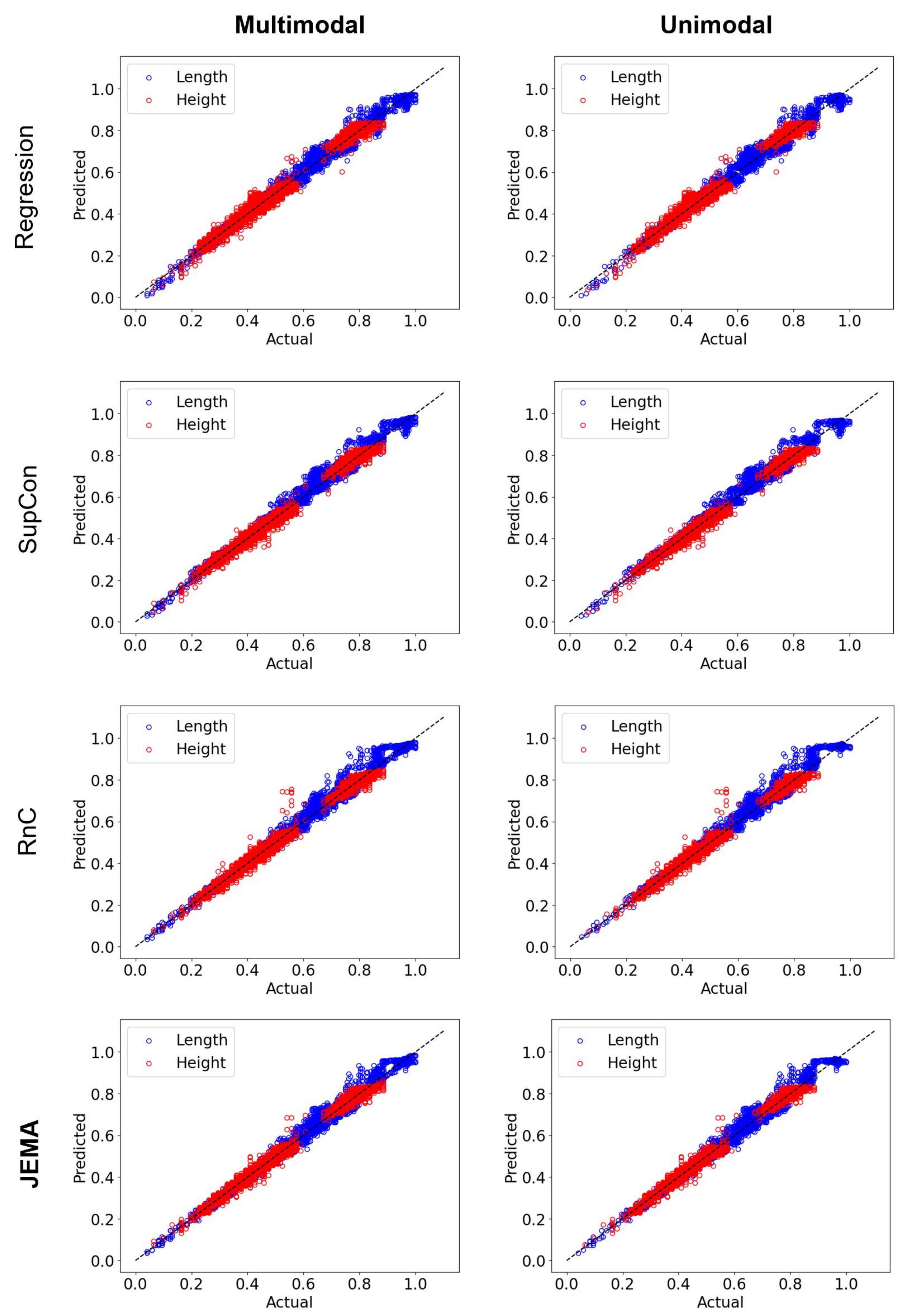}
\caption{Comparison of the melt pool length and height predictions in multi- and unimodal settings with different loss function approaches}
\label{slide_1}
\end{figure*}

% Clear the page and start the appendix
\clearpage

\section{Appendix B}
\label{ApxB}

\begin{figure*}[h!]
\centering
\includegraphics[width=0.990\textwidth]{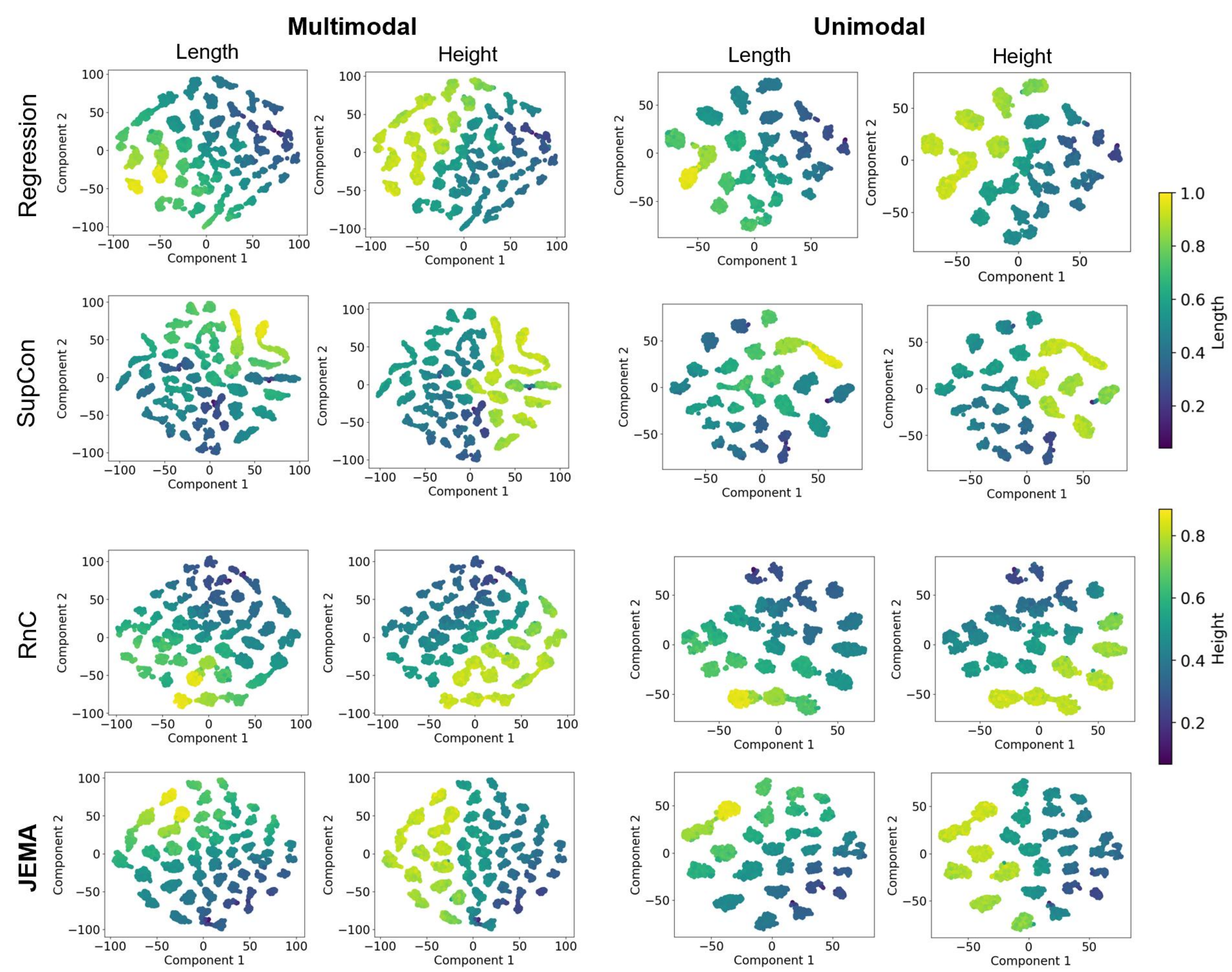}
\caption{Visual embedding representation using t-Distributed Stochastic Neighbor Embedding (t-SNE) with 2 components.}
\label{slide_2}
\end{figure*}

% Clear the page and start the appendix
\clearpage

\section{Appendix C}
\label{ApxC}

%%% IMAGEM %%%%%
\begin{figure*}[h!]
\centering
\subfloat[]{\includegraphics[width=0.45\textwidth]{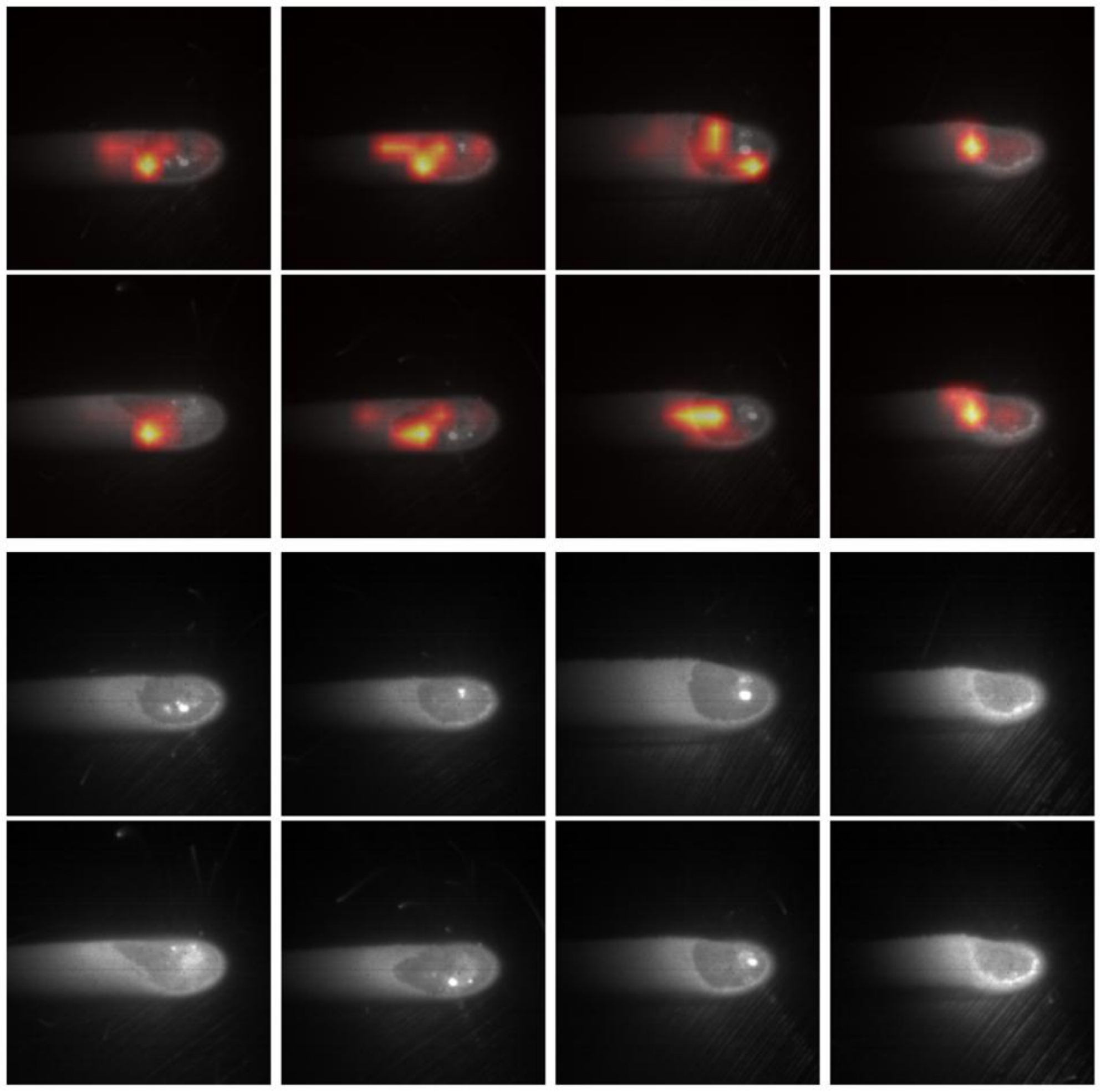}\label{attmap_1a}}
\hfil
\subfloat[]{\includegraphics[width=0.45\textwidth]{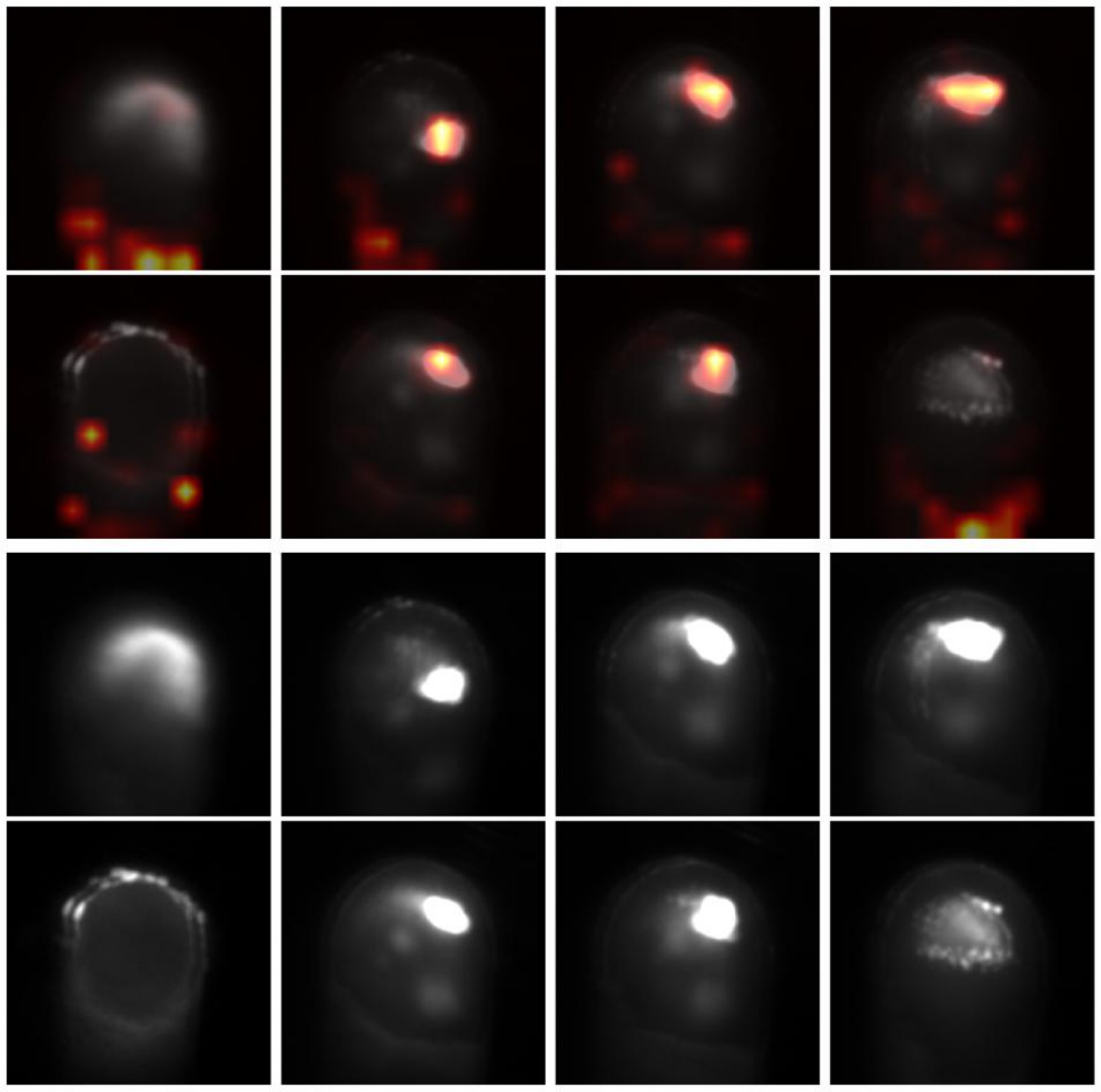}\label{attmap_1b}}

\caption{Attention maps visualization for (a) off-axis and (b) on-axis data, using JEMA cosie model.}
\label{fig_jema}
\end{figure*}

% Clear the page and start the appendix
\clearpage

\end{document}